\documentclass{article}

 \usepackage[preprint]{neurips_2025}

\usepackage[utf8]{inputenc} 
\usepackage[T1]{fontenc}    
\usepackage{hyperref}       
\usepackage{url}            

\usepackage{booktabs}       
\usepackage{amsfonts}       
\usepackage{nicefrac}       
\usepackage{microtype}      
\usepackage{xcolor}         
\usepackage{eurosym}
\usepackage{tikz}

\title{Computational Compliance for AI Regulation: Blueprint for a New Research Domain}

\author{%
  Bill Marino\thanks{Corresponding author.} \\
  University of Cambridge\\
  \texttt{wlm27@cam.ac.uk} \\
  \And
  Nicholas D. Lane \\
  University of Cambridge \\
}

\begin{document}

\maketitle

\begin{abstract}
The era of AI regulation (AIR) is upon us. But AI systems, we argue, will not be able to comply with these regulations at the necessary speed and scale by continuing to rely on traditional, analogue methods of compliance. Instead, we posit that compliance with these regulations will only realistically be achieved computationally: that is, with algorithms that run across the life cycle of an AI system, automatically steering it toward AIR compliance in the face of dynamic conditions. Yet despite their (we would argue) inevitability, the research community has yet to specify exactly how these algorithms for computational AIR compliance should behave --- or how we should benchmark their performance. To fill these gaps, we specify a set of design goals for such algorithms. In addition, we specify a benchmark dataset that could be used to quantitatively measure whether individual algorithms satisfy these design goals. By delivering this blueprint, we hope to give shape to an important but uncrystallized new domain of research --- and, in doing so, incite necessary investment in it. 
\end{abstract}

\section{Introduction}
\label{sec:intro}
This paper rests on the provocative premise that the future of all legal compliance is computational.

As every aspect of our lives becomes digitized, even if our laws are still printed in dust-gathering tomes and stenciled on road signs, compliance with those laws will be wholly managed by the architectures of --- and algorithms inside --- the digital systems that suffuse our world. 

The benefits of this computationally compliant future will be manifold. It will reduce the cost of compliance, removing a key barrier to markets and fostering competition \citep{KLAPPER2006591}. It will permit ``regulatory compliance in real time'' \citep{Bamidele2025IntegrationAIoT}, with violations mitigated as soon as they occur --- and, often, before any harm is done. What is more, by removing the potential for human error, computational compliance will ensure \textit{better} compliance, and a reality that hews closer to the letter of the laws that encode our societal values.    

As Artificial Intelligence Regulation (AIR) takes shape worldwide \citep{Alanoca_2025},
we argue that these regulations can (and should) represent the turning point in this evolution. ``Since AI is an algorithm,'' suggests one author, ``then the method of its regulation should be the use of an algorithm comprising legal standards'' \citep{Szostek2021IsTT}. 

In this paper, we sketch a blueprint for fulfilling that vision. In particular, we specify exactly how such an algorithm --- one that runs across the life cycle of an AI system, dynamically steering it towards AIR compliance in the face of variable conditions (e.g., post-deployment human feedback and data drift, changing legislation, and more) --- should behave. That is to say, we specify \textit{design goals} for computational AI regulation compliance (CAIRC). What is more, we specify a benchmarks that can be used to quantitatively measure progress toward many of those design goals. 

Above all, our hope is that this work brings structure and a set of lucid lodestars for future investment in this nascent but increasingly crucial field of research.


\section{Why Computational AIR Compliance Is Inevitable}
\label{sec:why}

\begin{quote} ``We built it, we trained it, but we don’t know what it’s doing.'' --- AI researcher \citep{Hassenfeld2023} \end{quote}

In short, we argue the expansiveness and expense of AI regulation is on a collision course with the complexity, scale, and dynamism of contemporary AI systems. The highly-manual \citep{Adams2025BusinessProcessCompliance} compliance methods of the past will prove unsustainable in this new reality, and CAIRC will emerge as the only feasible way for AI systems to comply with AIR. 

As mentioned, countries across the world are moving to regulate AI --- often with very different results \citep{sloane2025systematic, chun2024comparativeglobalairegulation, Alanoca_2025, doi:10.1177/03400352251384915}. If the European Union's Artificial Intelligence Act (EU AI Act) \citep{europa} --- dubbed ``the world’s first comprehensive AI law'' \citep{european_parliament_first} --- is any indication, then these regulations will sometimes have ``expansive scope'' \citep{addey2023charting} that reaches deep into the details of AI systems to dictate ``complex rules'' \citep{zulehner2024eu} around everything from their training data to their logging practices, and more \citep[Art. 10, 12]{europa}. 

Estimates suggest that relying on  traditional, human-driven methods to achieve compliance with these regulations will come at considerable expense to AI developers \citep{harvardComplianceCosts, Wagner2025, Haataja}\footnote{EU AI Act compliance costs for some types of AI systems, for example, are estimated to be as high as \euro{400,000} \citep[1872]{koh2024voices}.} 
By one estimate, the costs of complying with the EU AI Act alone could account for up to 17\%
of the total expense of an AI system \citep{laurer2021clarifying}. And of course many AI systems, increasingly aimed at global markets \citep{wto2024trading, reuters2025openai_cheapest_chatgpt}, will have to comply with multiple jurisdictions' AIRs, amplifying that percentage.

But we want to argue that, even were cost a non-issue, there is perhaps no amount of manual effort that could ever bring the AI systems of the future into AIR compliance at the necessary speed and scale. This is because, as AI systems grow larger, more complex, and more dynamic than ever before, their compliance ``surface area'' is rapidly outpacing what human compliance experts can feasibly tackle in a reasonable time frame. 

To wit, today's AI systems \citep{berkeleyShiftFrom} as well as the development pipelines \citep{Sadek2024} and supply chains behind them \citep{adalovelaceinstituteAllocatingAccountability, ceps, marino2024compliancecardsautomatedeu} have grown so complex that their own creators often struggle to understand them \citep{Hassenfeld2023}. These systems routinely comprise dozens of models, often externally sourced \citep{googleresearch, Renieris2023-ah, DBLP:journals/corr/abs-2405-13058, DBLP:journals/corr/abs-2406-08205, DBLP:conf/fat/LiesenfeldD24}. Their training sets, meanwhile, are nearing ``unimaginable scale'' \citep{codersstop2025_inconvenient_truth, shen2025llmsscalinghitwall, villalobos2024rundatalimitsllm}. As we consider a near future where AI systems include ``hundreds of agents'' \citep{falconer2025_morethanmachines}, this complexity and scale may continue rising. This, in turn, will make it increasingly impractical to rely on the contemporary norm \citep{FarleyLansang2025} of using human compliance experts to manually assess whether AI systems do or do not comply with a given AIR --- and using human AI developers to manually fix any compliance deficiencies identified during that assessment. Simply put, these new AI systems may go beyond what any one human --- or team of humans --- can meaningfully understand or manage without algorithmic assistance.\footnote{Note that this same notion of ``human impossible scale'' has also given rise to LLM validator functions that help ``scale [LLM] verification across benchmarks and tasks that would be infeasible for humans to manually check'' \citep{zhou2025variationverificationunderstandingverification}.} 

Adding fuel to the fire is the fact that modern AI systems ``are constantly changing and evolving" \citep{nicenboim2022_explanations} and, increasingly, the products of agile software development processes that favor continuous iteration \citep{balayn_gurses2024_misguided, carliniRapidIteration, DBLP:journals/corr/abs-2201-13224, DBLP:journals/tosem/Martinez-Fernandez22} and even of continual learning practices that perpetually update the system with inbound production data \citep{10444954}. This protean quality, especially when combined with the growing complexity and scale described above, will make it exceedingly difficult for time-consuming, human-led compliance protocols to maintain a compliant state in an AI system; as soon as these human operators conclude their assessment --- or render the relevant repairs --- the system is likely to have changed.  

These factors suggest that the human-driven regulatory compliance models of the past are destined to fail in the AIR setting
\citep{oreilly2025_eu_ai_embarrassment, krasadakis2023_regulate_ai, marino2024compliancecardsautomatedeu, lcfiActsTechnical, anderljung2023frontier, hacker2023regulating, siebel2024_ai_models_complex, 9659429}.
This will leave AI developers little choice but to shift to AIR compliance methods that are as scalable and dynamic as the AI systems themselves --- in other words, AIR compliance methods that are \textit{computational}.

\section{Deconstructing the problem}

\begin{quote} ``If you're overwhelmed by the whole, break it down into pieces.'' --- Chuck Close \citep{chuck_close_esq0102} \end{quote}

If, as we argue, computational AI regulation compliance (CAIRC) is inevitable, then how should its algorithms function? In other words, when developing them, what should our \textit{design goals} be? And, furthermore, how can we quantitatively measure progress toward those goals? 

To answer these questions, we find it useful to deconstruct CAIRC into two sub-problems. Specifically, we posit that any CAIRC algorithm must necessarily contain two complimentary functions, which we deem the \textit{Inspector} and the \textit{Mechanic}:\footnote{Happily, the \textit{Inspector} and \textit{Mechanic} have independent, standalone value. Even in the absence of a \textit{Mechanic} to automatically repair the compliance defects it identifies, the \textit{Inspector} can be used to alert human compliance assessors (or, if you will, ``human \textit{Mechanics}'') to compliance defects. Conversely, the \textit{Mechanic} can be used to automatically cure defects identified by human compliance assessors (or, if you will, ``human \textit{Inspectors}'').} 

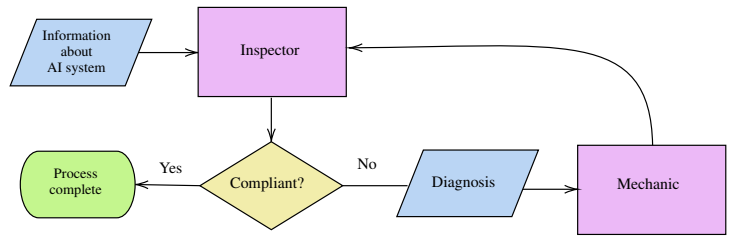
\begin{figure}[h!]
    \centering
\scalebox{0.6}{

\tikzset{every picture/.style={line width=0.75pt}} 

\begin{tikzpicture}[x=0.75pt,y=0.75pt,yscale=-1,xscale=1]

\draw    (409,280) -- (461,280) ;
\draw [shift={(463,280)}, rotate = 180] [color={rgb, 255:red, 0; green, 0; blue, 0 }  ][line width=0.75]    (10.93,-3.29) .. controls (6.95,-1.4) and (3.31,-0.3) .. (0,0) .. controls (3.31,0.3) and (6.95,1.4) .. (10.93,3.29)   ;
\draw  [fill={rgb, 255:red, 236; green, 185; blue, 247 }  ,fill opacity=1 ] (463,242) -- (587.5,242) -- (587.5,317) -- (463,317) -- cycle ;
\draw    (524.5,318) -- (524.5,353) ;
\draw [shift={(524.5,355)}, rotate = 270] [color={rgb, 255:red, 0; green, 0; blue, 0 }  ][line width=0.75]    (10.93,-3.29) .. controls (6.95,-1.4) and (3.31,-0.3) .. (0,0) .. controls (3.31,0.3) and (6.95,1.4) .. (10.93,3.29)   ;
\draw  [fill={rgb, 255:red, 236; green, 185; blue, 247 }  ,fill opacity=1 ] (782,358) -- (906.5,358) -- (906.5,433) -- (782,433) -- cycle ;
\draw  [fill={rgb, 255:red, 242; green, 237; blue, 171 }  ,fill opacity=1 ] (524.5,355) -- (584.5,392) -- (524.5,429) -- (464.5,392) -- cycle ;
\draw    (733.5,395) -- (780,395) ;
\draw [shift={(782,395)}, rotate = 180] [color={rgb, 255:red, 0; green, 0; blue, 0 }  ][line width=0.75]    (10.93,-3.29) .. controls (6.95,-1.4) and (3.31,-0.3) .. (0,0) .. controls (3.31,0.3) and (6.95,1.4) .. (10.93,3.29)   ;
\draw    (465.5,392) -- (412,391.04) ;
\draw [shift={(410,391)}, rotate = 1.03] [color={rgb, 255:red, 0; green, 0; blue, 0 }  ][line width=0.75]    (10.93,-3.29) .. controls (6.95,-1.4) and (3.31,-0.3) .. (0,0) .. controls (3.31,0.3) and (6.95,1.4) .. (10.93,3.29)   ;
\draw  [fill={rgb, 255:red, 196; green, 246; blue, 142 }  ,fill opacity=1 ] (328.94,363) -- (394.56,363) .. controls (403.09,363) and (410,375.54) .. (410,391) .. controls (410,406.46) and (403.09,419) .. (394.56,419) -- (328.94,419) .. controls (320.41,419) and (313.5,406.46) .. (313.5,391) .. controls (313.5,375.54) and (320.41,363) .. (328.94,363) -- cycle ;
\draw    (845,358) .. controls (843,258) and (769,275) .. (590,276) ;
\draw [shift={(590,276)}, rotate = 359.68] [color={rgb, 255:red, 0; green, 0; blue, 0 }  ][line width=0.75]    (10.93,-3.29) .. controls (6.95,-1.4) and (3.31,-0.3) .. (0,0) .. controls (3.31,0.3) and (6.95,1.4) .. (10.93,3.29)   ;
\draw  [fill={rgb, 255:red, 186; green, 213; blue, 244 }  ,fill opacity=1 ] (327.31,252) -- (424,252) -- (401.69,308) -- (305,308) -- cycle ;
\draw    (584.5,392) -- (639,392) ;
\draw  [fill={rgb, 255:red, 186; green, 213; blue, 244 }  ,fill opacity=1 ] (652.31,362) -- (749,362) -- (726.69,418) -- (630,418) -- cycle ;

\draw (492,272) node [anchor=north west][inner sep=0.75pt]   [align=left] {\begin{minipage}[lt]{44.68pt}\setlength\topsep{0pt}
\begin{center}
Inspector
\end{center}

\end{minipage}};
\draw (809,384) node [anchor=north west][inner sep=0.75pt]   [align=left] {\begin{minipage}[lt]{46.38pt}\setlength\topsep{0pt}
\begin{center}
Mechanic
\end{center}

\end{minipage}};
\draw (327,259.8) node [anchor=north west][inner sep=0.75pt]  [font=\small] [align=left] {\begin{minipage}[lt]{48.65pt}\setlength\topsep{0pt}
\begin{center}
Information\\about \\AI system
\end{center}

\end{minipage}};
\draw (483,383) node [anchor=north west][inner sep=0.75pt]   [align=left] {\begin{minipage}[lt]{54.32pt}\setlength\topsep{0pt}
\begin{center}
Compliant?
\end{center}

\end{minipage}};
\draw (593,368) node [anchor=north west][inner sep=0.75pt]   [align=left] {\begin{minipage}[lt]{15.76pt}\setlength\topsep{0pt}
\begin{center}
No
\end{center}

\end{minipage}};
\draw (426,371) node [anchor=north west][inner sep=0.75pt]   [align=left] {\begin{minipage}[lt]{19.36pt}\setlength\topsep{0pt}
\begin{center}
Yes
\end{center}

\end{minipage}};
\draw (332,376) node [anchor=north west][inner sep=0.75pt]  [font=\small] [align=left] {\begin{minipage}[lt]{39.97pt}\setlength\topsep{0pt}
\begin{center}
Process\\complete
\end{center}

\end{minipage}};
\draw (653,382.8) node [anchor=north west][inner sep=0.75pt]   [align=left] {\begin{minipage}[lt]{47.51pt}\setlength\topsep{0pt}
\begin{center}
Diagnosis
\end{center}

\end{minipage}};

\end{tikzpicture}

}
    \caption{\textbf{CAIRC flowchart}. As a first step, information about an AI system is submitted (e.g., by an overarching CAIRC algorithm) to an \textit{Inspector}. Next, the \textit{Inspector} reaches a finding of either compliance, in which case the process is complete (for the time being), or non-compliance, in which case the  \textit{Inspector} transmits its diagnosis to the \textit{Mechanic}. Upon receiving this diagnosis, the \textit{Mechanic} uses one or more automated tools to try to repair the diagnosed compliance defect(s). When finished, it calls the \textit{Inspector} to re-run its analysis. This loop repeats until the \textit{Inspector} finds that compliance exists, in which case the process has concluded (until, at least, it is triggered again).}
    \label{fig:levels}
\end{figure}

As depicted in Fig.~\ref{fig:levels}, the  \textit{Inspector} will diagnose --- at any given point in time and in a fully automated manner --- the AIR compliance level of an AI system.\footnote{In this way, one might say that the \textit{Inspector} plays a role similar to the human compliance assessors, for example, that feature prominently in the EU AI Act \citep[Art. 43]{europa}.} If the \textit{Inspector} finds that the AI system is non-compliant with one or more AIRs, it will communicate its diagnosis to the \textit{Mechanic}, which will endeavor to remedy the non-compliance using various automated tools, ultimately calling on the \textit{Inspector} to re-run its audit and determine if a compliance state has been achieved (or, perhaps, restored).    

In the sections that follow, we propose design goals and benchmarking methods for each of these two key functions --- the \textit{Inspector} and the \textit{Mechanic} --- as well as the overarching CAIRC algorithm that necessarily envelops them.

\vspace{1em}

\section{The \textit{Inspector}}
\label{sec:inspector}

In this section, we lay out the design goals (i.e., design criteria) for the CAIRC algorithm's \textit{Inspector} function. These relate to:
\begin{itemize}
    \item The \textit{Inspector}'s input;
    \item The \textit{Inspector}'s output; 
    \item The \textit{Inspector}'s internal function mapping the former to the latter.
\end{itemize}

Where applicable, we describe how close the state of the art (SOTA) comes to satisfying these design criteria and/or identify any open research problems that must be solved before these design criteria can realistically be achieved. In addition, we propose benchmarking methods for quantitatively assessing whether a given \textit{Inspector} algorithm satisfies these design criteria.

\subsection{Input}
\label{ssec:inspectorinput}

In order to assess the AIR compliance level of a given AI system, the \textit{Inspector} requires, as its input, information about that AI system. Importantly, this information --- and therefore the \textit{Inspector} input --- must satisfy the following design criteria: 

\paragraph{Comprehensive}: If an \textit{Inspector} is to accurately and holistically assess the AIR compliance level of an AI system, then the information inputted into it must describe \textit{all} aspects of the AI sytem that bear (or potentially bear) on AIR compliance. Failure to input all of the information relevant to AIR compliance carries great risk: specifically, of false positives (FP), whereby the \textit{Inspector} incorrectly labels a non-compliant AI system compliant because it is not privy to the facts evidencing otherwise. FPs like these will cause the \textit{Mechanic} to refrain from introducing necessary compliance-inducing repairs. This, in turn, could lead to penalties \citep[Art. 99]{europa} and even harm (of the sort the AIR aims to prevent). To mitigate this FP risk, \textit{Inspector} inputs must cover \textit{all} aspects of the AI that bear on its AIR compliance.  

So, for example, when it comes to the EU AI Act, the \textit{Inspector} input must include information about an AI system's data governance practices \citep[Art. 10]{europa}, human oversight mechanisms \citep[Art. 14]{europa}, and levels of accuracy \citep[Art. 15] {europa} --- all of which are the direct subjects of EU AI Act requirements. But it will also necessarily include information about that AI system's intended use, which determines the particular set of rules that apply to the AI system \citep[Art. 6]{europa}, and whether it is open source, which potentially exempts the AI system from those rules \citep[Art. 2]{europa}. The input to the \textit{Inspector} must therefore include the super set of all this information --- as well as any and all other information relevant to EU AI Act compliance. 
Oftentimes, this will represent a Brobdingnagian amount of data that is no small feat to assemble and feed to the \textit{Inspector}. 
What is more, this hurdle must be cleared for \textit{every AIR that the system is expected to comply with} --- potentially a heavy lift given the proliferous nature of AIR and the increasingly global footprint of AI systems \citep{wto2024trading, reuters2025openai_cheapest_chatgpt}.

\paragraph{Concurrent}: To avoid both FPs and false negatives (FN), it is also important that the \textit{Inspector} input reflect the current state of the AI system. In other words, the \textit{Inspector} must have up-to-date knowledge of all AIR-relevant facets of the system --- including transient ones like logs \citep[Art. 12]{europa}, anomalies \citep[Art. 14]{europa}, cyberattacks \citep[Art. 15]{europa}, and more. Information that is outdated --- even by a fraction of a second --- increases the risk of both FPs and FNs. As with comprehensiveness, while this design criteria appears technically feasible, the logistical challenge of achieving concurrency should not be underestimated.

\paragraph{Attestable}: Information that is relevant to an AI system's AIR compliance may have to be provided by untrusted sources. The EU AI Act, for example, includes a number of requirements around training data \citep[Art. 10]{europa}; in today's complex AI supply chain, this training data will often come from non-trusted providers via API or online communities like Hugging Face \citep{marino2024compliancecardsautomatedeu}.  In such cases, in order to avoid inaccuracies caused by either third party errors or attacks, it will be crucial to verify the externally-provided information is accurate \citep{lcfiActsTechnical, reuel2024openproblemstechnicalai}. Sometimes, due to confidentiality or other concerns, this will have to be accomplished without direct access to the subject of the verification (i.e., through ``remote attestation'' \citep{brundage2020trustworthyaidevelopmentmechanisms}). At the moment, this type of attestation is considered an ``open problem'' \citep{reuel2024openproblemstechnicalai}, though various methods are being explored
\citep{cen2024transparencyaccountabilitybackdiscussion, DBLP:journals/corr/abs-2402-02675, Sun2023PoT, huggingfaceFacebookbartlargecnnVerifyToken, schnabl2025attestableauditsverifiableai, scaramuzza2025showcomplyshowinganything}. 



\subsection{Output}
\label{ssec:inspectoroutput}

When the \textit{Inspector} finds that AIR compliance exists, it need not output anything other than, perhaps, a void return. In all other cases, the design criteria that the \textit{Inspector} output must satisfy is as follows:

\paragraph{\textit{Mechanic}-enabling}: The \textit{Inspector} output must provide enough information for the \textit{Mechanic} to fulfill its own role of repairing any identified compliance deficiencies in the AI system (i.e., it must be ``\textit{Mechanic}-enabling''). Among other things, this means that the \textit{Inspector} cannot, for example, simply return a binary class label of ``non-compliant'' or, differently, a single aggregate compliance score (like that of \citep{guldimann2024complaiframeworktechnicalinterpretation}). At a minimum, what is required are outputs that are granular enough that the \textit{Mechanic} knows what work to begin \textit{and where} --- without, in the interests of efficiency, needing to duplicate any of the compliance assessment work done by the \textit{Inspector}. For example, in communicating a violation of Article 10 of the EU AI Act, which regulates AI system training data, the \textit{Inspector} output would probably need to include, at minimum, a pointer to the non-compliant dataset along with the particular provision of Article 10 the dataset violates (let's say, for example, the provisions in Article 10, Section 3 requiring datasets to be reasonably free from errors). Given that, the \textit{Mechanic} should be sufficiently empowered to begin its work (e.g., of curing the excessive errors using probabilistic inference \citep{RekatsinasCIR17} or some other method).\footnote{Note that there may often be reason to keep some aspects of the AI out of the hands of the \textit{Inspector} --- for example, if the \textit{Inspector} is being operated by an arms-length auditor or a regulator (an arrangement would could have benefits in terms of providing an external check on the AI). In these situations, the \textit{Inspector} may not, by design, have access to enough information about the AI to provide a granular output to the \textit{Mechanic}.}  With anything less than this information, however, the \textit{Mechanic} would not be in a position to begin its work (at least, not without repeating the work of the \textit{Inspector}) --- and the output would therefore not be \textit{Mechanic}-enabling.


\subsection{Function mapping input to output}
\label{sssec:inspectoralgo}
The heart of the \textit{Inspector} is some function that maps its input onto its output; i.e., maps information about an AI system onto a \textit{Mechanic}-enabling AIR compliance diagnosis. This function might consist of an LLM  \citep{sovrano2025simplifying, li2025privacibenchevaluatingprivacycontextual, Makovec2024Preparing, Videsjorden2026, falconer2025_morethanmachines, tran}, a rule-based algorithm \citep{marino2024compliancecardsautomatedeu}, evaluation suites that run on the models or datasets comprising the AI system \citep{SOVRANO2023110866, DBLP:conf/wirtschaftsinformatik/WalkeBW23a, nolte2024robustness, buenomomcilovic2024assuring, DBLP:conf/emnlp/EsiobuTHUZFDPWS23, DBLP:journals/corr/abs-2308-03258, DBLP:conf/acl/LinHE22, DBLP:conf/acl/ParrishCNPPTHB22, guldimann2024complaiframeworktechnicalinterpretation, chen2024copybenchmeasuringliteralnonliteral}, or anything else. 
Regardless of this function's exact contents, it should satisfy the following design criteria: 
\paragraph{High accuracy, precision, and recall}: The \textit{Inspector}'s internal function must map inputs (information about AI systems) onto outputs (compliance predictions) with \textbf{high accuracy}. Because FPs (findings of compliance when an AI is, in fact, non-compliant) are especially costly in the AIR setting, it is critical for a CAIRC to have a low FP rate; i.e., \textbf{high precision}. That said, FNs can carry an undesirable cost as well: unnecessary --- and perhaps even compliance-jeopardizing --- repairs to an AI system that is, in fact, compliant. Therefore, \textbf{high recall}, while perhaps not as pressing as high precision, is also included as a design goal.
\paragraph{Low latency}: 
To avoid FPs and FNs, the \textit{Inspector} should perform its work with \textbf{low latency}. If it does not, given the dynamic nature of modern AI systems discussed in Sec. \ref{sec:intro}, the danger is that the AI system's actual compliance level has changed by the time the \textit{Inspector}'s internal function has finished executing, rendering its output erroneous.\footnote{Note that if latency approaches that of manual compliance analyses, then this potentially undermines some of the benefits of CAIRC put forth in Sec. \ref{sec:intro}).}

\section{The \textit{Mechanic}}
\label{sec:mechanic}

In this section, we lay out the design criteria for the CAIRC algorithm's \textit{Mechanic} function. These relate to:

\begin{itemize}
    \item The \textit{Mechanic}'s input;
    \item The \textit{Mechanics}'s output; 
    \item The compliance-inducing algorithm(s) employed by the \textit{Mechanic}.
\end{itemize}

Where applicable, we describe how close the SOTA comes to satisfying these design criteria and/or identify any open research problems that must be solved before these design criteria can realistically be satisfied. In addition, we propose benchmarking methods for quantitatively assessing whether a given \textit{Mechanic} algorithm satisfies these design criteria.  

\subsection{Input}
\label{ssec:mechanicinput}

The \textit{Mechanic} accepts, as its input, the \textit{Inspector} output. Thus the \textit{Inspector} output design criteria (Sec. \ref{ssec:inspectoroutput}) moonlight as the \textit{Mechanic} input design criteria. As previously discussed in Sec. \ref{ssec:inspectoroutput}, the granularity of this \textit{Inspector} output-cum-\textit{Mechanic} input may affect the necessary scope of the \textit{Mechanic}'s functionality and the details of its internal compliance-inducing algorithm.

\subsection{Output}
\label{ssec:mechanicoutput}

The \textit{Mechanic} is tasked with making compliance-inducing alterations to the AI system. This means directly editing the code, data, models, documentation, and other assets that compose the AI system. The core output of the \textit{Mechanic} is therefore the altered version of these assets (e.g., the datasets it has filtered, the models it has re-trained, the documentation it has amended, etc.).\footnote{In addition to the altered AI system, the \textit{Mechanic} should output a signal (e.g., a void function return) that indicates that its work, from its point of view, is complete. Upon receiving this signal, the overarching algorithm that encompasses the \textit{Mechanic} and the \textit{Inspector} can call on the \textit{Inspector} again, to check the \textit{Mechanic}'s work (i.e., to verify whether compliance has in fact been achieved or restored).} These altered AI assets (i.e., this output) should satisfy the following design criterion:

\paragraph{Deployable}: To reap the promised benefits of CAIRC, it is crucial that the altered AI system is readily deployable as-is. Here, our definition of deployability, borrowed from the world of software development, means amenable to automatic deployment \citep{HEYMANN202332, Schaefer2013ContinuousIntegration} ---- e.g., via continuous delivery \citep{Chen2015ContinuousDelivery}. Among other things, this means that the altered AI system assets, as outputted by the \textit{Mechanic}, must satisfy any computational or hardware constraints present in the production environment. This quality (and the use of automated deployment) is important because, if a human in the loop is required to test or deploy the amended AI system, this nibbles away at one of the core advantages of CAIRC discussed in Sec. \ref{sec:intro}: swift correction of compliance deficiencies, perhaps even before harm has occurred. 


\subsection{Compliance-inducing algorithm}
\label{sssec:mechanicalgo}

What lies between the input and the output of the \textit{Mechanic} is an algorithm that operates on the AI system in order to achieve or restore compliance. At a high level, we suggest this algorithm must have two key components: 
\begin{enumerate}
    \item A set of ``tools''
    or discrete functions that are called upon to repair specific compliance defects identified by the \textit{Inspector}. For instance, examples of tools a \textit{Mechanic} might wish to have at its disposal could include: 
    \begin{itemize}
        \item{Where non-compliance stems from biased (and unmitigated) outputs of a generative AI model \citep[Art. 9, 55]{europa}, a machine unlearning tool \citep{7163042, DBLP:journals/see/HineNTF24, DBLP:journals/tetci/XuWWJ24, marino2025bridgegapsmachineunlearning}, a model editing \citep{gupta2024unifiedframeworkmodelediting} tool, or a fine-tuning \citep{qi2023finetuningalignedlanguagemodels} tool, to try to reduce or eliminate the impact of the biased outputs --- without the need for full retraining of the model.}
        \item{Where non-compliance stems from model inaccuracy \citep[Art. 15]{europa}, tool(s) that let it improving accuracy by acquiring and then re-training on more or better data from new sources; this, in turn, may require the ability to generate synthetic data \citep{bauer2024comprehensiveexplorationsyntheticdata} or buy data on marketplaces --- as well as label, filter, or otherwise prepare that data for training, and, lastly, retrain and evaluate the downstream model.}
        \item{Where non-compliance stems from model leakage of personal data in the training set \citep[Art. 14]{europa}, a differential privacy (DP) tool \citep{bauer2024comprehensiveexplorationsyntheticdata, marino2025bridgegapsmachineunlearning}) that it can apply before retraining the model --- in order to mitigate the risk of leakage in the model;}
    \end{itemize}
    \item Some orchestrating algorithm that not only selects the right tools to use based on the contents of the \textit{Inspector}'s diagnosis, but manages the execution of those tools, monitors and navigates trade-offs, and, ultimately, makes a decision about when the amended AI system is compliant and can be outputted. 
 \end{enumerate}

Collectively, this algorithm and its component parts must satisfy the following design criteria: 

\paragraph{\textit{Inspector}-enabled}: Just as the \textit{Inspector} output must enable the \textit{Mechanic} to perform its work (Sec. \ref{ssec:inspectoroutput}, the \textit{Mechanic} algorithm must be able to finish the job and induce compliance given what the \textit{Inspector} has provided. Depending on the fidelity of the \textit{Inspector} outputs, this may sometimes warrant additional functionality in the \textit{Mechanic} (whether in its tools or its orchestrating algorithm). As an example, where an \textit{Inspector} output only shares that an Article 15 data poisoning violation (\citet[Art. 15]{europa}) has occurred, along with a dataset pointer, but does not share the particular data points that it suspects of being poisoned, the \textit{Mechanic} must possess the functionality to scan or analyze the AI system's datasets to identify those poisoned data points. Only then will it be able to being the work of mitigating them in order to restore compliance (e.g., by deleting them \citep{ibm_data_poisoning}). By contrast, where an \textit{Inspector} shares, in its output, a list of data points it suspects of being poisoned in violation of Article 15, the \textit{Mechanic} may potentially not require the same functionality.

\paragraph{Exhaustive}:  Importantly, to achieve true CAIRC, the \textit{Mechanic} algorithm must have access to a set of tools that, working together, can solve any arbitrary AIR compliance deficiency --- i.e., the set of tools is exhaustive.\footnote{To render their repairs, these tools must have the ability to edit the AI system: e.g., filter training sets, retrain models, and more. The \textit{Mechanic}, meanwhile, must possess the ability to map \textit{Inspector} outputs onto the relevant tools (e.g., through rule-based methods or by relying on an LLM to reason about which tools to leverage \citep{microsoft_agents_llms_2024}) --- and also to navigate trade-offs between different tool options based on features like their expected cost, latency, and their potential impact on model performance. The \textit{Mechanic} must also be able to orchestrate and manage the execution of those tools, through to some predicted state of completion. Once it has selected the specific tool(s) that it will use to address the non-compliance, the \textit{Mechanic} repair algorithm must orchestrate and manage the use of those tools to cure the particular deficiency. This includes the ability to monitor the progress and efficacy of these orchestrated tools -- i.e., as well as make a preliminary prediction about whether the tool has resolved the non-compliance (and, therefore, whether it is time to send an output message to the overarching algorithm that encompasses the \textit{Mechanic} and the \textit{Inspector}).} While there is work to be done mapping out the full spectrum of tools required by a \textit{Mechanic} to bring an AI system, under any scenario, back to a compliant state, it can be said with confidence that some of these tools, once identified, will not exist yet in the SOTA. In particular, we can assume that no tools yet exist wherever, in the eyes of scholars, AIR compliance calls for ``technical capabilities or engineering solutions that do not currently exist'' \citep{hairegulatory} or otherwise ``rest on open issues in computer science'' \citep{9659429}, including around transparency \citep{hairegulatory}, human oversight \citep{j4040043}, data quality \citep{j4040043, technologyreviewQuickGuide, microsoft, 9659429, microsoft, DBLP:journals/corr/abs-2409-00264}, and the robustness, explainability, and security of models  \citep{9659429, hairegulatory, technologyreviewQuickGuide, lcfiActsTechnical, Morley2020}. We revisit this topic in Future Work, \ref{sec:futurework}.

\paragraph{Trade-off-navigating}: Compliance will often come with trade-offs, including around cost \citep{dalli2021artificial} and performance \citep{kovari2024ai, Sanderson_2024}. The \textit{Mechanic} (most likely its orchestration algorithm) must not only select its tools so as to try to minimize these trade-offs, but must also continue to monitor these trade-offs as it leverages those tools, possibly abandoning some approaches when the trade-offs exceed some thresholds set by the AI developer (e.g., if machine unlearning performed on a  model to reduce bias ultimately degrades accuracy \citep{marino2025bridgegapsmachineunlearning} beyond acceptable limits). 

\paragraph{Effective}: Perhaps most importantly, the \textit{Mechanic} algorithm should have a high success rate at the task of restoring compliance given an arbitrary \textit{Inspector} diagnosis. Without this quality, we risk a situation where the \textit{Mechanic} enters an endless loop of attempting to restore compliance to an AI system, thinking it has done so, but ultimately having the \textit{Inspector} indicate it has failed at the task. Thusly, in parallel to looking for a high success rate, we should look for low rejection rates by the Inspection, number of attempts per success, and, generally speaking, low frequency of these types of loops.

\section{Connecting the \textit{Inspector} and \textit{Mechanic} in a closed-loop system}

The \textit{Inspector} and \textit{Mechanic} should ultimately be connected and encompassed by an overarching algorithm, creating a single, unified system for CAIRC. This closed-loop system will need to manage the following (Fig. \ref{fig:levels}): 

\begin{enumerate}
    \item Run the \textit{Inspector} routinely, perhaps as a scheduled job and ideally with enough frequency that AIR violations are detected and eliminated before harm is caused; 
     \item Route non-void \textit{Inspector} outputs (i.e., findings of non-compliance) to the \textit{Mechanic};
     \item When the \textit{Mechanic} returns, re-run the \textit{Inspector};
    \item Repeat this loop until the \textit{Inspector} returns void (indicating compliance has been restored);
\end{enumerate}

It is important to note that this unified system could, in theory, be split across multiple organizations. For example, the \textit{Mechanic} could be owned by an AI developer while the \textit{Inspector} could belong to an auditing company or even regulator. This would permit an external check on the compliance levels of the AI --- without given external entities access to certain (perhaps sensitive or confidential) parts of the AI system. 

The overarching CAIRC algorithm must also have the ability to detect an endless loop between the \textit{Mechanic} and the \textit{Inspector}, possibly triggering more severe mitigations, such as a pause of the AI system.


This closed loop system should satisfy all the design criteria of both the \textit{Inspector} and the \textit{Mechanic}.



 \section{Benchmark}

In this section, we outline a benchmark dataset that could be used to quantitatively assess progress towards many of the design goals we have set for the \textit{Inspector,} \textit{Mechanic}, and overarching CAIRC algorithm. When it comes to these types of algorithms, few attempts have been made to create benchmark datasets to measure their performance; moreover, those that do exist are strictly focused on LLM-based embodiments of \textit{Inspector}-type algorithms \citep{marino2025airegbenchbenchmarkinglanguagemodels, tran}.

To fill the void, the benchmark dataset that we propose would be composed of AI system ``snapshots.'' By ``snapshots,'' we mean that these samples would contain the full suite of assets comprising an AI system at a given point in time: its complete training and evaluation datasets, its model weights, its training, evaluation, and deployment code, its documentation and logs, etc. 

Importantly, some of the AI system snapshots in the dataset would be AIR-compliant, while others would be non-compliant; in the latter case, the snapshots would display diversity of non-compliance (e.g., violating different provisions of a given AIR). Those that are non-compliant would also be ``labeled'' with a ground-truth \textit{Inspector} output --- one that satisfies the design criteria laid out in Sec. \ref{ssec:inspectoroutput}, but is generated manually by human AIR compliance expert annotators.  

The harness accompanying this benchmark dataset, which would assist the various evaluations described below, would provide access to a \textit{Inspector} that is known to be robust and a \textit{Mechanic} that is known to be robust. 

So long as these ingredients are present, the benchmark dataset can be used to quantitatively assess the following:  

\subsection{Whether \textit{Inspector} outputs are \textit{Mechanic}-enabling}

To quantitatively assess whether a given \textit{Inspector}'s outputs are \textit{Mechanic}-enabling, non-compliant AI system snapshots could be inputted into that \textit{Inspector}. The corresponding \textit{Inspector} outputs could then be inputted, via the provided harness, into a \textit{Mechanic} that is known to be robust, to repair the identified defects. The AI system snapshot, once operated on by the \textit{Mechanic}, could then be run through a robust \textit{Inspector}, also via the harness, to determine whether the AI system has indeed achieved compliance in the wake of the repairs.\footnote{It is important to use a mature \textit{Inspector} for this assessment, lest we find ourselves in a loop whereby inadequate \textit{Mechanic} fixes are endorsed by an immature \textit{Inspector}.} In this scenarios, the \textit{Mechanic}'s success rate at making repairs would be a quantitative signal that the \textit{Inspector} outputs are indeed \textit{Mechanic}-enabling.

\subsection{The accuracy, precision, recall, and latency of the \textit{Inspector}'s internal function} 

To quantitatively measure the \textbf{accuracy}, \textbf{precision}, \textbf{recall}, and \textbf{latency} of the function that the \textit{Inspector} uses to map inputs onto outputs, AI snapshots (both compliant and non-compliant) can be inputted into the \textit{Inspector}. The corresponding outputs can then be compared to the ground truth \textit{Inspector} output ``labels'' in the benchmark dataset --- either by LLM-as-judge or numerous distance-based methods \citep{celikyilmaz2020evaluation, schmidtova2024automatic, wu2023large}. In addition to capturing the accuracy, precision, and recall with which the \textit{Inspector}'s internal function predicts the ground truth, we can capture the speed at which it does it.\footnote{Due to the potential subjectivity of assessing AIR compliant, the challenge of creating the ground truth for such a benchmark should not be underestimated; this is discussed in greater detail in Sec. \ref{sec:limitations}.} 

\subsection{Whether \textit{Mechanic} outputs are deployable} The evaluation harness that accompanies the benchmark dataset should set (or allow for the passing in of) parameters that represent deployment constraints (e.g., around model size or storage capacity). A \textit{Mechanic} should be evaluated for its ability to stay within these requirements when making alterations to an AI system snapshot. 

\subsection{Whether the \textit{Mechanic}'s repair algorithm is \textit{Inspector}-enable, comprehensive, effective, and trade-off navigating}
To quantitatively assess whether the \textit{Mechanic} repair algorithm satisfies various design criteria, including the ability to effectively repair AIR compliance defects, the \textit{Mechanic} algorithm should be fed \textit{Inspector} outputs from the benchmark dataset that indicate non-compliance and asked to operate on the associated AI system snapshot in order to repair the diagnosed compliance defect. The harness that accompanies the benchmark dataset should then give the \textit{Mechanic} access to a mature \textit{Inspector} to evaluate its repairs. In this manner, a \textit{Mechanic} repair algorithms could be evaluated for their success rate in achieving a compliant state, as graded by the mature \textit{Inspector} --- as well as the number of calls to the \textit{Inspector} required to induce compliance and their speed at doing so.\footnote{Note that measuring speed and cost, if possible, is also because it not only helps us compare \textit{Mechanic} algorithms, but helps us compare \textit{Mechanic} algorithms with human-driven compliance protocols. This might, in turn, support the hypothesis, put forth in Sec.~\ref{sec:intro}, that CAIRC can lower costs compared to human-driven compliance efforts.} The success rate of the \textit{Mechanic} could be segmented by type of compliance violation, to ensure exhaustive coverage of AIR defects. If the harness allows for the setting of trade-off constraints (e.g., around model accuracy or latency), this will help assess the ability of the \textit{Mechanic} to navigate various trade-offs when making its repairs.   

 \subsection{The effectiveness of the full closed-loop system}
 Although benchmarking the \textit{Inspector} and \textit{Mechanic} algorithms independently is valuable, it will ultimately be important to benchmark the tandem as well as the closed-loop CAIRC system that envelopes them. This will test the way they behave together, including how often they enter an endless loop and, working together, fail to cure a given AIR compliance deficiency. To assess this, non-compliant AI system snapshots from the benchmark can be inputted to a closed-loop system. The mature \textit{Inspector} available via the harness can be used as a model-as-judge \citep{gu2025surveyllmasajudge}, to assess the AIR compliance level of the AI system snapshot that remains after the CAIRC has run its full loop (i.e., after the \textit{Mechanic} has made its changes to the AI system and its colleague \textit{Inspector} has approved them). Alternatively, human AIR compliance experts could manually assess whether the resulting AI system snapshot is indeed compliant. Separately, the rate of failures (where the \textit{Inspector} and \textit{Mechanic} get caught in an endless loop) --- as well as the speed --- of the closed-loop system could be tracked. 

 \section{Limitations}
 \label{sec:limitations}

 In this section, we discuss known limitations of --- as well as anticipate critiques of --- the types of CAIRC algorithms we propose in this work:  

\subsection{The technical feasibility of AIR compliance and its measurement} 
Computationality aside, AIR compliance is haunted by existential questions about its technical feasibility and measurability \citep{guha2024ai, hairegulatory}. Critics argue that compliance with the EU AI Act, for example, rests on a number of open problems around explainability, human oversight, cybersecurity, and more \citep{guha2024ai, 9659429, hairegulatory, j4040043, technologyreviewQuickGuide, microsoft, 9659429, microsoft, DBLP:journals/corr/abs-2409-00264, technologyreviewQuickGuide, lcfiActsTechnical, Morley2020, lcfiActsTechnical}. Differently, it has been said that EU AI Act compliance will be difficult or even impossible to measure \citep{almada2023eu} due to a lack of agreed-upon benchmarks for core concepts like bias \citep{evans2020artificial, DBLP:journals/cacm/BuylB24, dulka2023use, Gornet2024} and interpretability \citep{hairegulatory, hutson2023rules}. Regarding LLMs in particular it has been said that it is ``impossible to demonstrate compliance with a given regulatory specification'' \citep{10.1093/polsoc/puae020, SAEED2023110273, Donghyeok}. These critiques foreshadow potential hurdles en route to CAIRC, of course. Because if researchers have not yet figured out, using any method, how to measure or achieve AIR compliance in certain scenarios, how can we expect our \textit{Inspector} and \textit{Mechanic} to do so? 

\subsection{The subjectivity of compliance}
As a separate matter, when it comes to compliance, there are those that hold the viewpoint that ``[h]uman oversight, nuanced judgment, ethical considerations, and strategic thinking cannot, and should not, be outsourced entirely to algorithms'' \citep{compliance2025_future_compliance}. This may stem from the notion that compliance, in general, is ``hard to measure'' and ``not binary'' \citep{wu2021compliance}. Needless to say, making AIR compliance computational (and especially benchmarking it) requires the opposite view: that compliance can successfully be encoded in digital systems that make, in some cases, binary predictions (e.g., compliant or not-compliant) --- with their performance quantitatively measured using objective ground truth. If AIR compliance ``gray areas'' truly exist, then this jeopardizes the value and viability of CAIRC. Accordingly, it is a potential feature of this domain worth monitoring closely as we develop CAIRC algorithms. 

\section{Future Work}
\label{sec:futurework}

In this paper, we have striven to create a scaffold for future research on the topic of computational AIR compliance --- a scaffold that we invite the research community to fill in. That said, within this scaffold, here are some especially salient or pressing areas of future work that we wish to highlight: 

\subsection{Remote attestation of \textit{Inspector} inputs} As discussed in Sec. \ref{ssec:inspectorinput}, amid an increasingly complex AI supply chain where AI systems are likely to include multiple third party models and datasets \citep{marino2024compliancecardsautomatedeu}, including from untrusted providers and sometimes behind APIs, the remote attestation of these AI system components (and, in particular, the aspects of them that bear on AIR compliance) remains an open problem \citep{reuel2024openproblemstechnicalai}. Additional work in this area will help create a foundation of trust upon which CAIRC can be built. 

\subsection{Multi-modal \textit{Inspectors}}
Work has been done thus to create algorithms that automatically assess the AIR compliance of an AI system; but, thus far, this work suffers from blind spots. This work has included LLMs that assess the AIR compliance of an AI system based strictly on text artifacts such as technical documentation  \citep{sovrano2025simplifying, li2025privacibenchevaluatingprivacycontextual, Makovec2024Preparing, Videsjorden2026, falconer2025_morethanmachines, tran}, transparency artifacts \citep{marino2024compliancecardsautomatedeu} or logs \citep{Videsjorden2026}. It has also included AIR-specific evaluation suites that strictly evaluate the models or datasets comprising the AI system \citep{SOVRANO2023110866, DBLP:conf/wirtschaftsinformatik/WalkeBW23a, nolte2024robustness, buenomomcilovic2024assuring, DBLP:conf/emnlp/EsiobuTHUZFDPWS23, DBLP:journals/corr/abs-2308-03258, DBLP:conf/acl/LinHE22, DBLP:conf/acl/ParrishCNPPTHB22, guldimann2024complaiframeworktechnicalinterpretation, chen2024copybenchmeasuringliteralnonliteral}. What is still left to do, however, is to combine these approaches and develop \textit{Inspector} algorithms that can scrutinize \textit{all} aspects of an AI system (its models, datasets, text artifacts, and everything else available) in order to render a more comprehensive AIR compliance analysis.  

\subsection{Mapping out a full set of \textit{Mechanic} tools}

There is work to be done mapping out the full spectrum of tools required by the \textit{Mechanic} to bring the AI system, under any scenario, back to a compliant state. A sampling of such tools can be seen in Sec. \ref{sssec:mechanicalgo}, but we believe this list represents a mere fraction of the necessary set. A full taxonomy of such tools will help guide and prioritize their development by the research community.

\subsection{The open problems of AIR compliance}

This line of work has a dependency on solving the open problems that continue to surround AIR compliance \citep{hairegulatory, 9659429}. This includes open problems around transparency \citep{hairegulatory}, human oversight \citep{j4040043}, data quality \citep{j4040043, technologyreviewQuickGuide, microsoft, 9659429, microsoft, DBLP:journals/corr/abs-2409-00264}, and ensuring the robustness, explainability, and security of models  \citep{9659429, hairegulatory, technologyreviewQuickGuide, lcfiActsTechnical, Morley2020}. It also includes open problems related to the measurement of AIR compliance, especially as regards LLMs and other frontier AI systems \citep{almada2023eu, evans2020artificial, DBLP:journals/cacm/BuylB24, dulka2023use, Gornet2024, hairegulatory, hutson2023rules, 10.1093/polsoc/puae020, SAEED2023110273, Donghyeok}.

\section{Conclusion}

Legal compliance, we argue, will ultimately be governed not by human oversight but by algorithms operating within digital systems. AI regulation represents a prime opportunity to begin the transition to this future of computational compliance. To move the field forward, we propose a set of design principles to steer the development of computational AIR compliance algorithms and, additionally, sketch a benchmark to quantitatively measure how well algorithms satisfy the design principles. Our intention in laying out this framework is to help coalesce an important new research area that is still being formed --- and to spark additional research investment in it.

{
\small

\bibliographystyle{plainnat}
\bibliography{references}

@misc{marino2024compliancecardsautomatedeu,
      title={Compliance {C}ards: Automated {EU} {AI} {A}ct Compliance Analyses amidst a Complex {AI} Supply Chain}, 
      author={Bill Marino and Yaqub Chaudhary and Yulu Pi and Rui-Jie Yew and Preslav Aleksandrov and Carwyn Rahman and William F. Shen and Isaac Robinson and Nicholas D. Lane},
      year={2024},
      eprint={2406.14758},
      archivePrefix={arXiv},
      primaryClass={cs.AI},
      url={https://arxiv.org/abs/2406.14758}, 
}

@legislation{europa,
    title = {{Artificial Intelligence Act}},
    shorttitle = {EU AI Act},
    author={{European Union}},
    organization={European Union},
    type = {Regulation},
    year = {2024},
    month = {March},
    day = {13},
    location = {Brussels, Belgium},
    note = {Official Journal of the European Union},
    url = {https://eur-lex.europa.eu/legal-content/EN/TXT/?uri=CELEX:52021PC0206}
}

@misc{berkeleyShiftFrom,
	author = {Matei Zaharia and Omar Khattab and Lingjiao Chen and Jared Quincy Davis and Heather Miller and Chris Potts and James Zou and Michael Carbin and Jonathan Frankle and Naveen Rao and Ali Ghodsi},
	title = {The Shift from Models to Compound {A}{I} Systems},
	howpublished = {\url{https://bair.berkeley.edu/blog/2024/02/18/compound-ai-systems/}},
	year = {2024},
	note = {[Accessed 22-10-2025]},
}

@misc{adalovelaceinstituteAllocatingAccountability,
	author = {Ian Brown},
	title = {{A}llocating accountability in {A}{I} supply chains},
	howpublished = {\url{https://www.adalovelaceinstitute.org/resource/ai-supply-chains/}},
	year = {2023},
	note = {[Accessed 22-10-2025]},
}

@misc{ceps,
	author = {Alex Engler and Andrea Renda},
	title = {Reconciling the {A}{I} Value Chain with the {E}{U}’s {A}rtificial {I}ntelligence {A}ct},
	howpublished = {\url{https://www.ceps.eu/ceps-publications/reconciling-the-ai-value-chain-with-the-eus-artificial-intelligence-act/}},
	year = {2022},
	note = {[Accessed 22-10-2025]},
}

@MISC{googleresearch,

  author       = {Shamik Chaudhuri and Kingshuk Dasgupta and Isaac Hepworth, Michael Le and Mark Lodato and Mihai Maruseac and Sarah Meiklejohn and Tehila Minkus and Kara Olive},
  title        = "Securing the {AI} software supply chain",
  howpublished = "\url{https://research.google/pubs/securing-the-ai-software-supply-chain/}",
	note = {[Accessed 22-08-2025]},
  language     = "en",
  year    =  {2024}
}

@ARTICLE{Renieris2023-ah,
  title   = "Building Robust {RAI} Programs as Third-Party {AI} tools
             proliferate",
  author  = {Elizabeth M. Renieris and David Kiron and Steven Mills},
  journal = "MIT Sloan Manage. Rev",
url = {https://sloanreview.mit.edu/projects/building-robust-rai-programs-as-third-party-ai-tools-proliferate/},
  year    =  {2023}
}

@article{DBLP:journals/corr/abs-2405-13058,
  author       = {Cailean Osborne and
                  Jennifer Ding and
                  Hannah Rose Kirk},
  title        = {The {AI} Community Building the Future? {A} Quantitative Analysis
                  of Development Activity on {H}ugging {F}ace {H}ub},
  journal      = {CoRR},
  volume       = {abs/2405.13058},
  year         = {2024},
  url          = {https://doi.org/10.48550/arXiv.2405.13058},
  doi          = {10.48550/ARXIV.2405.13058},
  eprinttype    = {arXiv},
  eprint       = {2405.13058},
  bibsource    = {dblp computer science bibliography, https://dblp.org}
}

@article{DBLP:journals/corr/abs-2406-08205,
  author       = {Jason Jones and
                  Wenxin Jiang and
                  Nicholas Synovic and
                  George K. Thiruvathukal and
                  James C. Davis},
  title        = {What do we know about {H}ugging {F}ace? {A} systematic literature review
                  and quantitative validation of qualitative claims},
  journal      = {CoRR},
  volume       = {abs/2406.08205},
  year         = {2024},
  url          = {https://doi.org/10.48550/arXiv.2406.08205},
  doi          = {10.48550/ARXIV.2406.08205},
  eprinttype    = {arXiv},
  eprint       = {2406.08205},
  bibsource    = {dblp computer science bibliography, https://dblp.org}
}

@inproceedings{DBLP:conf/fat/LiesenfeldD24,
  author       = {Andreas Liesenfeld and
                  Mark Dingemanse},
  title        = {Rethinking open source generative {AI:} {O}pen washing and the {EU}
                  {AI} {A}ct},
  booktitle    = {The 2024 {ACM} Conference on Fairness, Accountability, and Transparency,
                  FAccT 2024, Rio de Janeiro, Brazil, June 3-6, 2024},
  pages        = {1774--1787},
  publisher    = {{ACM}},
  year         = {2024},
  url          = {https://doi.org/10.1145/3630106.3659005},
  doi          = {10.1145/3630106.3659005},
  bibsource    = {dblp computer science bibliography, https://dblp.org}
}

@inproceedings{Videsjorden2026,
  author    = {Videsjorden, Adela Nedisan and Nikolov, Nikolay and Lien, Carl-Henrik and Goknil, Arda and Sen, Sagar and Song, Hui and Soylu, Ahmet and Roman, Dumitru},
  title     = {Positioning {LLM}-Enabled Agents as Legal Compliance Aides for Data Pipelines},
  booktitle = {Rules and Reasoning. RuleML+RR 2025},
  series    = {Lecture Notes in Computer Science},
  volume    = {16144},
  pages     = {227--236},
  publisher = {Springer, Cham},
  year      = {2026},
  doi       = {10.1007/978-3-032-08887-1\_14},
  isbn      = {978-3-032-08887-1}
}

@article{RekatsinasCIR17,
  author    = {Theodoros Rekatsinas and Xu Chu and Ihab F. Ilyas and Christopher R{\'e}},
  title     = {Holo{C}lean: Holistic Data Repairs with Probabilistic Inference},
  journal   = {Proceedings of the VLDB Endowment},
  volume    = {10},
  number    = {11},
  pages     = {1190--1201},
  year      = {2017},
  doi       = {10.14778/3137628.3137631},
  url       = {https://www.vldb.org/pvldb/vol10/p1190-rekatsinas.pdf}
}

@article{FarleyLansang2025,
  title        = {{AI} Auditing: First Steps Towards the Effective Regulation of Artificial Intelligence Systems},
  author       = {Farley, Edwin A. and Lansang, Christian R.},
  journal      = {Harvard Journal of Law \& Technology},
  volume       = {38},
  number       = {Digest},
  pages        = {--},
  year         = {2025},
  url          = {https://jolt.law.harvard.edu/assets/digestImages/Farley-Lansang-AI-Auditing-publication-2.13.2025.pdf},
  note         = {Accessed: 2025-12-28}
}

@misc{carliniRapidIteration,
	author = {Nicholas Carlini},
	title = {Rapid Iteration in Machine Learning Research},
	howpublished = {\url{https://nicholas.carlini.com/writing/2022/rapid-iteration-machine-learning-research.html}},
	year = {2022},
	note = {[Accessed 22-10-2025]},
}

@article{DBLP:journals/corr/abs-2201-13224,
  author       = {David Piorkowski and
                  John T. Richards and
                  Michael Hind},
  title        = {Evaluating a Methodology for Increasing {AI} Transparency: {A} Case
                  Study},
  journal      = {CoRR},
  volume       = {abs/2201.13224},
  year         = {2022},
  url          = {https://arxiv.org/abs/2201.13224},
  eprinttype    = {arXiv},
  eprint       = {2201.13224},
  bibsource    = {dblp computer science bibliography, https://dblp.org}
}

@online{ibm_data_poisoning,
  author       = {Tom Krantz and Alexandra Jonker},
  title        = {What Is Data Poisoning?},
  year         = {2025},
  url          = {https://www.ibm.com/think/topics/data-poisoning},
  organization = {IBM},
  note         = {Accessed: 2026-01-01}
}

@article{HEYMANN202332,
title = {Assessment Framework for Deployability of Machine Learning Models in Production},
journal = {Procedia CIRP},
volume = {118},
pages = {32-37},
year = {2023},
note = {16th CIRP Conference on Intelligent Computation in Manufacturing Engineering},
issn = {2212-8271},
doi = {https://doi.org/10.1016/j.procir.2023.06.007},
url = {https://www.sciencedirect.com/science/article/pii/S2212827123002299},
author = {Henrik Heymann and Hendrik Mende and Maik Frye and Robert H. Schmitt}
}

@incollection{Schaefer2013ContinuousIntegration,
  author       = {Schäfer, Andreas and Reichenbach, Marc and Fey, Dietmar},
  title        = {Continuous Integration and Automation for {D}ev{O}ps},
  booktitle    = {IAENG Transactions on Engineering Technologies: Special Edition of the World Congress on Engineering and Computer Science 2011},
  editor       = {Kim, Haeng Kon and Ao, S.-I. and Rieger, Burghard B.},
  series       = {Lecture Notes in Electrical Engineering},
  volume       = {170},
  publisher    = {Springer Netherlands},
  address      = {Dordrecht, The Netherlands},
  year         = {2013},
  pages        = {345--358},
  isbn         = {9400747853},
  doi          = {10.1007/978-94-007-4786-9_28},
}

@article{Chen2015ContinuousDelivery,
  author    = {Lianping Chen},
  title     = {Continuous Delivery: Huge Benefits, but Challenges Too},
  journal   = {IEEE Software},
  volume    = {32},
  number    = {2},
  pages     = {50--54},
  year      = {2015},
  doi       = {10.1109/MS.2015.27},
  url       = {https://ieeexplore.ieee.org/document/7006384/}
}

@techreport{dalli2021artificial,
  author       = {Hubert Dalli},
  title        = {Artificial {I}ntelligence {A}ct: Initial Appraisal of a {E}uropean {C}ommission Impact Assessment},
  institution  = {European Parliamentary Research Service (EPRS), European Parliament},
  type         = {EPRS Briefing},
  number       = {PE 694.212},
  year         = {2021},
  month        = {July},
  url          = {https://www.europarl.europa.eu/RegData/etudes/BRIE/2021/694212/EPRS_BRI(2021)694212_EN.pdf},
  note         = {Accessed: 2025-12-29}
}

@inproceedings{Sanderson_2024,
  author       = {Conrad Sanderson and Emma Schleiger and David M. Douglas and Petra Kuhnert and Qinghua Lu},
  title        = {Resolving Ethics Trade-offs in Implementing Responsible {AI}},
  booktitle    = {Proceedings of the IEEE Conference on Artificial Intelligence (CAI 2024)},
  year         = {2024},
  doi          = {10.1109/CAI59869.2024.00215},
  url          = {https://www.researchgate.net/publication/382732079_Resolving_Ethics_Trade-offs_in_Implementing_Responsible_AI},
  publisher    = {IEEE},
  note         = {Also available as arXiv preprint arXiv:2401.08103}
}

@article{kovari2024ai,
  author    = {Attila Kovari},
  title     = {{AI} for Decision Support: Balancing Accuracy, Transparency, and Trust Across Sectors},
  journal   = {Information},
  year      = {2024},
  volume    = {15},
  number    = {11},
  pages     = {725},
  doi       = {10.3390/info15110725},
  url       = {https://www.mdpi.com/2078-2489/15/11/725},
  publisher = {MDPI},
  note      = {Open access},
}

@INPROCEEDINGS{tran,
  author={Tran, Quynh and Salg, Josef and Shpileuskaya, Krystsina and Wang, Qi and Putzar, Larissa and Blankenburg, Sven},
  booktitle={2025 International Joint Conference on Neural Networks (IJCNN)}, 
  title={Bridging {AI} and Regulation: Large Language Models for Documentation Compliance Check}, 
  year={2025},
  volume={},
  number={},
  pages={1-10},
  doi={10.1109/IJCNN64981.2025.11229064}}

@misc{lcfiActsTechnical,
	author = {Bill Marino},
	title = {The {E}{U} {A}{I} {A}ct’s Technical “Tension Areas”},
	howpublished = {\url{https://www.lcfi.ac.uk/news-events/blog/post/the-eu-ai-acts-technical-tension-areas}},
	year = {2024},
	note = {[Accessed 22-10-2025]},
}

@misc{harvardComplianceCosts,
	author = {Weiyue Wu and Shaoshan Liu},
	title = {{W}hy Compliance Costs of {A}{I} Commercialization May Be Holding Start-Ups Back},
	howpublished = {\url{https://studentreview.hks.harvard.edu/why-compliance-costs-of-ai-commercialization-maybe-holding-start-ups-back/}},
	year = {2023},
	note = {[Accessed 22-10-2025]},
}

@misc{laurer2021clarifying,
    title={Clarifying the costs for the {EU}'s {AI} Act},
    author={Laurer, Moritz and Renda, Andrea and Yeung, Timothy},
    year={2021},
    month={September},
    day={24},
    howpublished={Centre for European Policy Studies},
    url={https://www.ceps.eu/clarifying-the-costs-for-the-eus-ai-act/},
    type={Policy Brief},
    organization={CEPS}
}

@TechReport{Haataja,
  author={Haataja, Meeri and Bryson, Joanna J.},
  title={{What costs should we expect from the EU’s AI Act?}},
  year=2021,
  month=Aug,
  institution={Center for Open Science},
  type={SocArXiv},
  url={https://ideas.repec.org/p/osf/socarx/8nzb4.html},
  number={8nzb4},
  doi={10.31219/osf.io/8nzb4},
}

@article{Adams2025BusinessProcessCompliance,
  author  = {Adams, N. and Augusto, A. and Davern, M. and et al.},
  title   = {Addressing the Contemporary Challenges of Business Process Compliance},
  journal = {Business \& Information Systems Engineering},
  year    = {2025},
  doi     = {10.1007/s12599-025-00929-3},
  note    = {Online ahead of print}
}

@article{Hassenfeld2023,
  author = {Hassenfeld, Noam},
  title = {Even the scientists who build {AI} can’t tell you how it works},
  journal = {Vox – Unexplainable Podcast},
  year = {2023},
  month = {Jul 15},
  url = {https://www.vox.com/unexplainable/2023/7/15/23793840/chat-gpt-ai-science-mystery-unexplainable-podcast},
  note = {Accessed: 2025-11-30}
}

@Article{Morley2020,
author={Morley, Jessica
and Floridi, Luciano
and Kinsey, Libby
and Elhalal, Anat},
title={From What to How: {A}n Initial Review of Publicly Available {AI} Ethics Tools, Methods and Research to Translate Principles into Practices},
journal={Science and Engineering Ethics},
year={2020},
month={Aug},
day={01},
volume={26},
number={4},
pages={2141-2168},
issn={1471-5546},
doi={10.1007/s11948-019-00165-5},
url={https://doi.org/10.1007/s11948-019-00165-5}
}

@Article{Sadek2024,
author={Sadek, Malak
and Kallina, Emma
and Bohn{\'e}, Thomas
and Mougenot, C{\'e}line
and Calvo, Rafael A.
and Cave, Stephen},
title={Challenges of responsible {AI} in practice: {S}coping review and recommended actions},
journal={AI {\&} SOCIETY},
year={2024},
month={Feb},
day={19},
issn={1435-5655},
doi={10.1007/s00146-024-01880-9},
url={https://doi.org/10.1007/s00146-024-01880-9}
}

@misc{technologyreviewQuickGuide,
	author = {Melissa Heikkilä},
	title = {{A} quick guide to the most important {A}{I} law you’ve never heard of},
	howpublished = {\url{https://www.technologyreview.com/2022/05/13/1052223/guide-ai-act-europe/}},
	year = {2022},
	note = {[Accessed 22-08-2025]},
}

@article{DBLP:journals/corr/abs-2409-00264,
  author       = {Nuno Sousa e Silva},
  title        = {The {A}rtificial {I}ntelligence {A}ct: {C}ritical overview},
  journal      = {CoRR},
  volume       = {abs/2409.00264},
  year         = {2024},
  url          = {https://doi.org/10.48550/arXiv.2409.00264},
  doi          = {10.48550/ARXIV.2409.00264},
  eprinttype    = {arXiv},
  eprint       = {2409.00264},
  bibsource    = {dblp computer science bibliography, https://dblp.org}
}

@INPROCEEDINGS{7163042,
  author={Yinzhi Cao and Junfeng Yang},
  booktitle={2015 IEEE Symposium on Security and Privacy}, 
  title={Towards Making Systems Forget with Machine Unlearning}, 
  year={2015},
  volume={},
  number={},
  pages={463-480},
  doi={10.1109/SP.2015.35}}

@misc{bauer2024comprehensiveexplorationsyntheticdata,
      title={Comprehensive Exploration of Synthetic Data Generation: A Survey}, 
      author={André Bauer and Simon Trapp and Michael Stenger and Robert Leppich and Samuel Kounev and Mark Leznik and Kyle Chard and Ian Foster},
      year={2024},
      eprint={2401.02524},
      archivePrefix={arXiv},
      primaryClass={cs.LG},
      url={https://arxiv.org/abs/2401.02524}, 
}

@article{DBLP:journals/tetci/XuWWJ24,
  author       = {Jie Xu and
                  Zihan Wu and
                  Cong Wang and
                  Xiaohua Jia},
  title        = {Machine Unlearning: Solutions and Challenges},
  journal      = {{IEEE} Trans. Emerg. Top. Comput. Intell.},
  volume       = {8},
  number       = {3},
  pages        = {2150--2168},
  year         = {2024},
  url          = {https://doi.org/10.1109/TETCI.2024.3379240},
  doi          = {10.1109/TETCI.2024.3379240},
  bibsource    = {dblp computer science bibliography, https://dblp.org}
}

@misc{hairegulatory,
	author = {Neel Guha and Christie M. Lawrence and Lindsey A. Gailmard and Kit T. Rodolfa and Faiz Surani and Rishi Bommasani and
Inioluwa Deborah Raji and Mariano-Florentino Cuéllar and
Colleen Honigsberg and Percy Liang and Daniel E. Ho},
	title = {The {AI} Regulatory
Alignment Problem},
	howpublished = {\url{https://hai.stanford.edu/sites/default/files/2023-11/AI-Regulatory-Alignment.pdf}},
	year = {},
	note = {[Accessed 22-08-2025]},
}

@techreport{wto2024trading,
  author       = {World Trade Organization},
  title        = {Trading with Intelligence: How {AI} Shapes and Is Shaped by International Trade},
  institution  = {World Trade Organization},
  year         = {2024},
  month        = {nov},
  type         = {Report},
  url          = {https://www.wto.org/english/res_e/booksp_e/trading_with_intelligence_e.pdf},
  note         = {Comprehensive WTO Secretariat report on artificial intelligence and international trade}
}

@techreport{zulehner2024eu,
  author       = {Zulehner, Bruno},
  title        = {{EU Artificial Intelligence Act}: Regulating the Use of Facial Recognition Technologies in Publicly Accessible Spaces},
  institution  = {Stanford-Vienna Transatlantic Technology Law Forum, European Union Law Working Paper No.\ 91},
  year         = {2024},
  note         = {European Union Law Working Papers, edited by Siegfried Fina and Roland Vogl},
  url          = {https://law.stanford.edu/wp-content/uploads/2024/06/EU-Law-WP-91-Zulehner.pdf}
}

@misc{microsoft,
	author = {Microsoft},
	title = {Microsoft’s Response to the {E}uropean
{C}ommission’s Consultation on the
{A}rtificial {I}ntelligence {A}ct},
	howpublished = {\url{https://blogs.microsoft.com/wp-content/uploads/prod/sites/73/2021/09/microsoft-response-to-the-european-commission-consultation-on-the-artifical-intelligence-act.pdf}},
	year = {2021},
	note = {[Accessed 22-08-2025]},
}

@inproceedings{koh2024voices,
    author = {Koh, Florence and Grosse, Kathrin and Apruzzese, Giovanni},
    title = {Voices from the Frontline: Revealing the {AI} Practitioners' viewpoint on the {European AI Act}},
    booktitle = {Proceedings of the Hawaii International Conference on System Sciences},
    series = {HICSS},
    year = {2024},
    location = {Maui, Hawaii}
}

@techreport{almada2023eu,
    author = {Almada, Marco and Petit, Nicolas},
    title = {The {EU} {AI} Act: A Medley of Product Safety and Fundamental Rights?},
    institution = {European University Institute},
    type = {Working Paper},
    series = {RSC},
    number = {2023/59},
    year = {2023},
    url = {https://hdl.handle.net/1814/75982}
}

@misc{zhou2025variationverificationunderstandingverification,
      title={Variation in Verification: Understanding Verification Dynamics in Large Language Models}, 
      author={Yefan Zhou and Austin Xu and Yilun Zhou and Janvijay Singh and Jiang Gui and Shafiq Joty},
      year={2025},
      eprint={2509.17995},
      archivePrefix={arXiv},
      primaryClass={cs.CL},
      url={https://arxiv.org/abs/2509.17995}, 
}

@article{hutson2023rules,
    author = {Hutson, Matthew},
    title = {Rules to keep {AI} in check: nations carve different paths for tech regulation},
    journal = {Nature},
    year = {2023},
    month = {August},
    volume = {620},
    number = {7973},
    pages = {260--263},
    doi = {10.1038/d41586-023-02491-y},
    note = {PMID: 37553464}
}

@INPROCEEDINGS{9659429,
  author={Fiazza, Maria-Camilla},
  booktitle={2021 20th International Conference on Advanced Robotics (ICAR)}, 
  title={The {EU} Proposal for Regulating {AI}: Foreseeable Impact on Medical Robotics}, 
  year={2021},
  volume={},
  number={},
  pages={222-227},
  doi={10.1109/ICAR53236.2021.9659429}}

@Article{j4040043,
AUTHOR = {Ebers, Martin and Hoch, Veronica R. S. and Rosenkranz, Frank and Ruschemeier, Hannah and Steinrötter, Björn},
TITLE = {The {E}uropean {C}ommission’s Proposal for an {A}rtificial {I}ntelligence {A}ct—A Critical Assessment by Members of the {R}obotics and {AI} {L}aw {S}ociety ({RAILS})},
JOURNAL = {J},
VOLUME = {4},
YEAR = {2021},
NUMBER = {4},
PAGES = {589--603},
URL = {https://www.mdpi.com/2571-8800/4/4/43},
ISSN = {2571-8800},
DOI = {10.3390/j4040043}
}

@misc{brundage2020trustworthyaidevelopmentmechanisms,
      title={Toward Trustworthy {AI} Development: Mechanisms for Supporting Verifiable Claims}, 
      author={Miles Brundage and Shahar Avin and Jasmine Wang and Haydn Belfield and Gretchen Krueger and Gillian Hadfield and Heidy Khlaaf and Jingying Yang and Helen Toner and Ruth Fong and Tegan Maharaj and Pang Wei Koh and Sara Hooker and Jade Leung and Andrew Trask and Emma Bluemke and Jonathan Lebensold and Cullen O'Keefe and Mark Koren and Théo Ryffel and JB Rubinovitz and Tamay Besiroglu and Federica Carugati and Jack Clark and Peter Eckersley and Sarah de Haas and Maritza Johnson and Ben Laurie and Alex Ingerman and Igor Krawczuk and Amanda Askell and Rosario Cammarota and Andrew Lohn and David Krueger and Charlotte Stix and Peter Henderson and Logan Graham and Carina Prunkl and Bianca Martin and Elizabeth Seger and Noa Zilberman and Seán Ó hÉigeartaigh and Frens Kroeger and Girish Sastry and Rebecca Kagan and Adrian Weller and Brian Tse and Elizabeth Barnes and Allan Dafoe and Paul Scharre and Ariel Herbert-Voss and Martijn Rasser and Shagun Sodhani and Carrick Flynn and Thomas Krendl Gilbert and Lisa Dyer and Saif Khan and Yoshua Bengio and Markus Anderljung},
      year={2020},
      eprint={2004.07213},
      archivePrefix={arXiv},
      primaryClass={cs.CY},
      url={https://arxiv.org/abs/2004.07213}, 
}

@article{DBLP:journals/tosem/Martinez-Fernandez22,
  author       = {Silverio Mart{\'{\i}}nez{-}Fern{\'{a}}ndez and
                  Justus Bogner and
                  Xavier Franch and
                  Marc Oriol and
                  Julien Siebert and
                  Adam Trendowicz and
                  Anna Maria Vollmer and
                  Stefan Wagner},
  title        = {Software Engineering for {AI}-Based Systems: {A} Survey},
  journal      = {{ACM} Trans. Softw. Eng. Methodol.},
  volume       = {31},
  number       = {2},
  pages        = {37e:1--37e:59},
  year         = {2022},
  url          = {https://doi.org/10.1145/3487043},
  doi          = {10.1145/3487043},
  bibsource    = {dblp computer science bibliography, https://dblp.org}
}

@inproceedings{DBLP:conf/acl/ParrishCNPPTHB22,
  author       = {Alicia Parrish and
                  Angelica Chen and
                  Nikita Nangia and
                  Vishakh Padmakumar and
                  Jason Phang and
                  Jana Thompson and
                  Phu Mon Htut and
                  Samuel R. Bowman},
  editor       = {Smaranda Muresan and
                  Preslav Nakov and
                  Aline Villavicencio},
  title        = {{BBQ:} {A} hand-built bias benchmark for question answering},
  booktitle    = {Findings of the Association for Computational Linguistics: {ACL} 2022,
                  Dublin, Ireland, May 22-27, 2022},
  pages        = {2086--2105},
  publisher    = {Association for Computational Linguistics},
  year         = {2022},
  url          = {https://doi.org/10.18653/v1/2022.findings-acl.165},
  doi          = {10.18653/V1/2022.FINDINGS-ACL.165},
  bibsource    = {dblp computer science bibliography, https://dblp.org}
}

@inproceedings{DBLP:conf/acl/LinHE22,
  author       = {Stephanie Lin and
                  Jacob Hilton and
                  Owain Evans},
  editor       = {Smaranda Muresan and
                  Preslav Nakov and
                  Aline Villavicencio},
  title        = {{TruthfulQA}: Measuring How Models Mimic Human Falsehoods},
  booktitle    = {Proceedings of the 60th Annual Meeting of the Association for Computational
                  Linguistics (Volume 1: Long Papers), {ACL} 2022, Dublin, Ireland,
                  May 22-27, 2022},
  pages        = {3214--3252},
  publisher    = {Association for Computational Linguistics},
  year         = {2022},
  url          = {https://doi.org/10.18653/v1/2022.acl-long.229},
  doi          = {10.18653/V1/2022.ACL-LONG.229},
  bibsource    = {dblp computer science bibliography, https://dblp.org}
}

@misc{reuel2024openproblemstechnicalai,
      title={Open Problems in Technical {AI} Governance}, 
      author={Anka Reuel and Ben Bucknall and Stephen Casper and Tim Fist and Lisa Soder and Onni Aarne and Lewis Hammond and Lujain Ibrahim and Alan Chan and Peter Wills and Markus Anderljung and Ben Garfinkel and Lennart Heim and Andrew Trask and Gabriel Mukobi and Rylan Schaeffer and Mauricio Baker and Sara Hooker and Irene Solaiman and Alexandra Sasha Luccioni and Nitarshan Rajkumar and Nicolas Moës and Jeffrey Ladish and Neel Guha and Jessica Newman and Yoshua Bengio and Tobin South and Alex Pentland and Sanmi Koyejo and Mykel J. Kochenderfer and Robert Trager},
      year={2024},
      eprint={2407.14981},
      archivePrefix={arXiv},
      primaryClass={cs.CY},
      url={https://arxiv.org/abs/2407.14981}, 
}

@article{DBLP:journals/corr/abs-2402-02675,
  author       = {Tobin South and
                  Alexander Camuto and
                  Shrey Jain and
                  Shayla Nguyen and
                  Robert Mahari and
                  Christian Paquin and
                  Jason Morton and
                  Alex 'Sandy' Pentland},
  title        = {Verifiable evaluations of machine learning models using {Z}k{SNARK}s},
  journal      = {CoRR},
  volume       = {abs/2402.02675},
  year         = {2024},
  url          = {https://doi.org/10.48550/arXiv.2402.02675},
  doi          = {10.48550/ARXIV.2402.02675},
  eprinttype    = {arXiv},
  eprint       = {2402.02675},
  bibsource    = {dblp computer science bibliography, https://dblp.org}
}

@misc{huggingfaceFacebookbartlargecnnVerifyToken,
	author = {{Hugging Face}},
	title = {{A}dd verify{T}oken field to verify evaluation results are produced by {H}ugging {F}ace's automatic model evaluator},
	howpublished = {\url{https://huggingface.co/facebook/bart-large-cnn/discussions/23}},
	year = {2024},
	note = {[Accessed 22-10-2025]},
}

@techreport{evans2020artificial,
    title = {Artificial Intelligence and Public Standards: A Review by the {Committee} on {Standards} in {Public Life}},
    author = {{Committee on Standards in Public Life}},
    year = {2020},
    month = {February},
    institution = {Government of the United Kingdom},
    type = {Government Review},
    note = {Chair: Lord Evans of Weardale KCB DL},
    url = {https://assets.publishing.service.gov.uk/media/5e553b3486650c10ec300a0c/Web_Version_AI_and_Public_Standards.PDF}
}

@article{DBLP:journals/cacm/BuylB24,
  author       = {Maarten Buyl and
                  Tijl De Bie},
  title        = {Inherent Limitations of {AI} Fairness},
  journal      = {Commun. {ACM}},
  volume       = {67},
  number       = {2},
  pages        = {48--55},
  year         = {2024},
  url          = {https://doi.org/10.1145/3624700},
  doi          = {10.1145/3624700},
  bibsource    = {dblp computer science bibliography, https://dblp.org}
}

@article{dulka2023use,
    author = {Dulka, Anne},
    title = {The Use of Artificial Intelligence in International Human Rights Law},
    journal = {Stanford Technology Law Review},
    volume = {26},
    pages = {316},
    year = {2023},
    shortjournal = {Stan. Tech. L. Rev.}
}

@misc{guldimann2024complaiframeworktechnicalinterpretation,
      title={{COMPL-AI} Framework: A Technical Interpretation and {LLM} Benchmarking Suite for the {EU} {A}rtificial {I}ntelligence {A}ct}, 
      author={Philipp Guldimann and Alexander Spiridonov and Robin Staab and Nikola Jovanović and Mark Vero and Velko Vechev and Anna Gueorguieva and Mislav Balunović and Nikola Konstantinov and Pavol Bielik and Petar Tsankov and Martin Vechev},
      year={2024},
      eprint={2410.07959},
      archivePrefix={arXiv},
      primaryClass={cs.CL},
      url={https://arxiv.org/abs/2410.07959}, 
}

@article{DBLP:journals/see/HineNTF24,
  author       = {Emmie Hine and
                  Claudio Novelli and
                  Mariarosaria Taddeo and
                  Luciano Floridi},
  title        = {Supporting Trustworthy {AI} Through Machine Unlearning},
  journal      = {Sci. Eng. Ethics},
  volume       = {30},
  number       = {5},
  pages        = {43},
  year         = {2024},
  url          = {https://doi.org/10.1007/s11948-024-00500-5},
  doi          = {10.1007/S11948-024-00500-5},
  bibsource    = {dblp computer science bibliography, https://dblp.org}
}

@article{SOVRANO2023110866,
title = {An objective metric for Explainable {AI}: How and why to estimate the degree of explainability},
journal = {Knowledge-Based Systems},
volume = {278},
pages = {110866},
year = {2023},
issn = {0950-7051},
doi = {https://doi.org/10.1016/j.knosys.2023.110866},
url = {https://www.sciencedirect.com/science/article/pii/S0950705123006160},
author = {Francesco Sovrano and Fabio Vitali}
}

@inproceedings{DBLP:conf/wirtschaftsinformatik/WalkeBW23a,
  author       = {Fabian Walke and
                  Lars Bennek and
                  Till J. Winkler},
  title        = {Artificial Intelligence Explainability Requirements of the {AI} {A}ct
                  and Metrics for Measuring Compliance},
  booktitle    = {Digital Responsibility: Social, Ethical, Ecological Implications of
                  IS, 18. Internationale Tagung Wirtschaftsinformatik {(WI} 2023), September
                  18-21, 2023, Paderborn, Germany},
  pages        = {77},
  publisher    = {AISeL},
  year         = {2023},
  url          = {https://aisel.aisnet.org/wi2023/77},
  bibsource    = {dblp computer science bibliography, https://dblp.org}
}

@article{nolte2024robustness,
   title={Robustness and Cybersecurity in the {EU} {A}rtificial {I}ntelligence {A}ct},
   author={Nolte, Henrik and Rateike, Miriam and Finck, Michele},
   year={2024},
   url={https://blog.genlaw.org/pdfs/genlaw_icml2024/4.pdf}
}

@misc{cen2024transparencyaccountabilitybackdiscussion,
      title={From Transparency to Accountability and Back: A Discussion of Access and Evidence in {AI} Auditing}, 
      author={Sarah H. Cen and Rohan Alur},
      year={2024},
      eprint={2410.04772},
      archivePrefix={arXiv},
      primaryClass={cs.CY},
      url={https://arxiv.org/abs/2410.04772}, 
}

@inproceedings{DBLP:conf/emnlp/EsiobuTHUZFDPWS23,
  author       = {David Esiobu and
                  Xiaoqing Ellen Tan and
                  Saghar Hosseini and
                  Megan Ung and
                  Yuchen Zhang and
                  Jude Fernandes and
                  Jane Dwivedi{-}Yu and
                  Eleonora Presani and
                  Adina Williams and
                  Eric Michael Smith},
  editor       = {Houda Bouamor and
                  Juan Pino and
                  Kalika Bali},
  title        = {{ROBBIE:} {R}obust Bias Evaluation of Large Generative Language Models},
  booktitle    = {Proceedings of the 2023 Conference on Empirical Methods in Natural
                  Language Processing, {EMNLP} 2023, Singapore, December 6-10, 2023},
  pages        = {3764--3814},
  publisher    = {Association for Computational Linguistics},
  year         = {2023},
  url          = {https://doi.org/10.18653/v1/2023.emnlp-main.230},
  doi          = {10.18653/V1/2023.EMNLP-MAIN.230},
  bibsource    = {dblp computer science bibliography, https://dblp.org}
}

@article{DBLP:journals/corr/abs-2308-03258,
  author       = {Tianrui Qin and
                  Xitong Gao and
                  Juanjuan Zhao and
                  Kejiang Ye and
                  Cheng{-}Zhong Xu},
  title        = {{APBench}: {A} Unified Benchmark for Availability Poisoning Attacks
                  and Defenses},
  journal      = {CoRR},
  volume       = {abs/2308.03258},
  year         = {2023},
  url          = {https://doi.org/10.48550/arXiv.2308.03258},
  doi          = {10.48550/ARXIV.2308.03258},
  eprinttype    = {arXiv},
  eprint       = {2308.03258},
  bibsource    = {dblp computer science bibliography, https://dblp.org}
}

@inproceedings{Makovec2024Preparing,
    author={Makovec, Barbara and Rei, Luis and Novalija, Inna},
    title={Preparing {AI} for Compliance: Initial Steps of a Framework for Teaching {LLMs} to Reason About Compliance},
    booktitle={Companion Proceedings of the 8th International Joint Conference on Rules and Reasoning (RuleML+RR'24)},
    year={2024},
    month={September},
    address={Bucharest, Romania},
    publisher={CEUR Workshop Proceedings},
    volume={3816},
    url={https://ceur-ws.org/Vol-3816/paper63.pdf}
}

@misc{chun2024comparativeglobalairegulation,
      title={Comparative Global {AI} Regulation: Policy Perspectives from the {EU}, {C}hina, and the {US}}, 
      author={Jon Chun and Christian Schroeder de Witt and Katherine Elkins},
      year={2024},
      eprint={2410.21279},
      archivePrefix={arXiv},
      primaryClass={cs.CY},
      url={https://arxiv.org/abs/2410.21279}, 
}

@inproceedings{Sun2023PoT,
    author={Sun, Haochen and Zhang, Hongyang},
    title={{PoT}: Securely Proving Legitimacy of Training Data and Logic for {AI} Regulation},
    booktitle={ICML 2023 Workshop on Generative AI and Law},
    year={2023},
    url={https://blog.genlaw.org/CameraReady/22.pdf}
}

@article{SAEED2023110273,
title = {Explainable {AI} ({XAI}): A systematic meta-survey of current challenges and future opportunities},
journal = {Knowledge-Based Systems},
volume = {263},
pages = {110273},
year = {2023},
issn = {0950-7051},
doi = {https://doi.org/10.1016/j.knosys.2023.110273},
url = {https://www.sciencedirect.com/science/article/pii/S0950705123000230},
author = {Waddah Saeed and Christian Omlin}
}

@article{10.1093/polsoc/puae020,
    author = {Judge, Brian and Nitzberg, Mark and Russell, Stuart},
    title = {When code isn’t law: rethinking regulation for artificial intelligence},
    journal = {Policy and Society},
    pages = {puae020},
    year = {2024},
    month = {05},
    issn = {1449-4035},
    doi = {10.1093/polsoc/puae020},
    url = {https://doi.org/10.1093/polsoc/puae020},
    eprint = {https://academic.oup.com/policyandsociety/advance-article-pdf/doi/10.1093/polsoc/puae020/57986099/puae020.pdf},
}

@article{KLAPPER2006591,
title = {Entry regulation as a barrier to entrepreneurship},
journal = {Journal of Financial Economics},
volume = {82},
number = {3},
pages = {591-629},
year = {2006},
issn = {0304-405X},
doi = {https://doi.org/10.1016/j.jfineco.2005.09.006},
url = {https://www.sciencedirect.com/science/article/pii/S0304405X06000936},
author = {Leora Klapper and Luc Laeven and Raghuram Rajan}
}

@inproceedings{buenomomcilovic2024assuring,
  title={Assuring Compliance of {LLM}s with {EU} {AIA} Robustness Demands},
  author={Momcilovic, Tomas Bueno and Buesser, Beat and Zizzo, Giulio and Purcell, Mark and Balta, Dian},
  booktitle={Wirtschaftsinformatik 2024 Proceedings},
  year={2024},
  pages={126},
  url={https://aisel.aisnet.org/wi2024/126}
}

@article{Szostek2021IsTT,
  title={Is the Traditional Method of Regulation (the Legislative Act) Sufficient to Regulate Artificial Intelligence, or Should It Also Be Regulated by an Algorithmic Code?},
  author={Dariusz Szostek},
  journal={Białostockie Studia Prawnicze},
  year={2021},
  volume={26},
  pages={43 - 60},
  url={https://api.semanticscholar.org/CorpusID:239476730}
}

@techreport{Gornet2024,
  author = {Gornet, Mélanie},
  title = {The {AI Act}: the evolution of "trustworthy {AI}" from policy documents to mandatory regulation},
  year = {2024},
  note = {ffhal-04785519f}
}

@article{microsoft_agents_llms_2024,
    title = {How agents and copilots work with {LLM}s},
    author = {{Microsoft}},
    journal = {Microsoft Learn},
    year = {2024},
    month = {November},
    day = {24},
    url = {https://learn.microsoft.com/en-us/dotnet/ai/conceptual/agents}
}

@misc{chen2024copybenchmeasuringliteralnonliteral,
      title={{CopyBench}: Measuring Literal and Non-Literal Reproduction of Copyright-Protected Text in Language Model Generation}, 
      author={Tong Chen and Akari Asai and Niloofar Mireshghallah and Sewon Min and James Grimmelmann and Yejin Choi and Hannaneh Hajishirzi and Luke Zettlemoyer and Pang Wei Koh},
      year={2024},
      eprint={2407.07087},
      archivePrefix={arXiv},
      primaryClass={cs.CL},
      url={https://arxiv.org/abs/2407.07087}, 
}

@inproceedings{Donghyeok,
author = {Lee, Donghyeok and Todorova, Christina and Dehghani, Alireza},
year = {2024},
month = {12},
pages = {41-46},
title = {Ethical Risks and Future Direction in Building Trust for Large Language Models Application under the {EU AI Act}},
doi = {10.1145/3701268.3701272}
}

@article{guha2024ai,
    title={{AI} Regulation Has Its Own Alignment Problem: The Technical and Institutional Feasibility of Disclosure, Registration, Licensing, and Auditing},
    author={Guha, Neel and Lawrence, Christie M. and Gailmard, Lindsey A. and Rodolfa, Kit T. and Surani, Faiz and Bommasani, Rishi and Raji, Inioluwa Deborah and Cu{\'e}llar, Mariano-Florentino and Honigsberg, Colleen and Liang, Percy and Ho, Daniel E.},
    journal={George Washington Law Review},
    volume={92},
    number={6},
    pages={1473},
    year={2024},
    publisher={The George Washington University Law Review}
}

@article{european_parliament_first,
    title = {{EU AI Act}: First Regulation on Artificial Intelligence},
    author = {{European Parliament}},
    journal = {European Parliament Topics},
    year = {2024},
    month = {June},
    day = {18},
    url = {https://www.europarl.europa.eu/topics/en/article/20230601STO93804/eu-ai-act-first-regulation-on-artificial-intelligence},
    urldate = {2024-06-18}
}

@misc{anderljung2023frontier,
    title={Frontier {AI} Regulation: Managing Emerging Risks to Public Safety},
    author={Anderljung, Markus and Barnhart, Emma and Korinek, Anton and Leung, Jeffrey and O'Keefe, Cullen and Whittlestone, Jess and others},
    year={2023},
    howpublished={Unpublished manuscript}
}

@inproceedings{hacker2023regulating,
    title={Regulating {ChatGPT} and other Large Generative {AI} Models},
    author={Hacker, Philipp and Engel, Andreas and Mauer, Marco},
    booktitle={2023 ACM Conference on Fairness, Accountability, and Transparency},
    series={FAccT '23},
    year={2023},
    pages={14},
    location={Chicago, IL, USA},
    publisher={Association for Computing Machinery},
    address={New York, NY, USA},
    doi={10.1145/3593013.3594067}
}

@misc{qi2023finetuningalignedlanguagemodels,
      title={Fine-tuning Aligned Language Models Compromises Safety, Even When Users Do Not Intend To!}, 
      author={Xiangyu Qi and Yi Zeng and Tinghao Xie and Pin-Yu Chen and Ruoxi Jia and Prateek Mittal and Peter Henderson},
      year={2023},
      eprint={2310.03693},
      archivePrefix={arXiv},
      primaryClass={cs.CL},
      url={https://arxiv.org/abs/2310.03693}, 
}

@article{wu2021compliance,
   title={Compliance Dynamism: Capturing the Polynormative and Situational Nature of Business Responses to Law},
   author={Wu, Yixin and van Rooij, Benjamin},
   journal={Journal of Business Ethics},
   volume={168},
   pages={579--591},
   year={2021},
   publisher={Springer},
   doi={10.1007/s10551-019-04234-4}
}

@inproceedings{Alanoca_2025, series={FAccT ’25},
   title={Comparing Apples to Oranges: A Taxonomy for Navigating the Global Landscape of {AI} Regulation},
   url={http://dx.doi.org/10.1145/3715275.3732059},
   DOI={10.1145/3715275.3732059},
   booktitle={Proceedings of the 2025 ACM Conference on Fairness, Accountability, and Transparency},
   publisher={ACM},
   author={Alanoca, Sacha and Gur-Arieh, Shira and Zick, Tom and Klyman, Kevin},
   year={2025},
   month=jun, pages={914–937},
   collection={FAccT ’25} }

@misc{marino2025airegbenchbenchmarkinglanguagemodels,
      title={{AIReg-Bench}: Benchmarking Language Models That Assess {AI} Regulation Compliance}, 
      author={Bill Marino and Rosco Hunter and Zubair Jamali and Marinos Emmanouil Kalpakos and Mudra Kashyap and Isaiah Hinton and Alexa Hanson and Maahum Nazir and Christoph Schnabl and Felix Steffek and Hongkai Wen and Nicholas D. Lane},
      year={2025},
      eprint={2510.01474},
      archivePrefix={arXiv},
      primaryClass={cs.AI},
      url={https://arxiv.org/abs/2510.01474}, 
}

@misc{li2025privacibenchevaluatingprivacycontextual,
      title={{PrivaCI-Bench}: Evaluating Privacy with Contextual Integrity and Legal Compliance}, 
      author={Haoran Li and Wenbin Hu and Huihao Jing and Yulin Chen and Qi Hu and Sirui Han and Tianshu Chu and Peizhao Hu and Yangqiu Song},
      year={2025},
      eprint={2502.17041},
      archivePrefix={arXiv},
      primaryClass={cs.CL},
      url={https://arxiv.org/abs/2502.17041}, 
}

@article{sloane2025systematic,
  title={A systematic review of regulatory strategies and transparency mandates in {AI} regulation in {E}urope, the {U}nited {S}tates, and {C}anada},
  author={Sloane, Mona and W{\"u}llhorst, Elena},
  journal={Data \& Policy},
  volume={7},
  pages={e11},
  year={2025},
  publisher={Cambridge University Press}
}

@article{Bamidele2025IntegrationAIoT,
  author       = {Bamidele, Matthew},
  title        = {Integration of {AI} with {IoT} for Real-Time Compliance in Connected Insurance},
  journal      = {ResearchGate},
  year         = {2025},
  month        = {August},
  day          = {18},
  note         = {Uploaded 20 August 2025},
  url          = {https://www.researchgate.net/publication/394753646\_Integration\_of\_AI\_with\_IoT\_for\_Real-Time\_Compliance\_in\_Connected\_Insurance}
}

@online{falconer2025_morethanmachines,
  author       = {Sean Falconer},
  title        = {More Than Machines: The Inner Workings of {AI} Agents},
  year         = {2025},
  month        = mar,
  day          = {5},
  journal      = {Medium},
  url          = {https://seanfalconer.medium.com/more-than-machines-the-inner-workings-of-ai-agents-5bba7904d04e}
}

@misc{gupta2024unifiedframeworkmodelediting,
      title={A Unified Framework for Model Editing}, 
      author={Akshat Gupta and Dev Sajnani and Gopala Anumanchipalli},
      year={2024},
      eprint={2403.14236},
      archivePrefix={arXiv},
      primaryClass={cs.LG},
      url={https://arxiv.org/abs/2403.14236}, 
}

@online{compliance2025_future_compliance,
  author       = {{Compliance Podcast Network}},
  title        = {Stepping Up and Stepping Forward: The Future of Compliance in an Age of {AI} and Deregulation},
  year         = {2025},
  month        = apr,
  day          = {4},
  journal      = {Compliance Podcast Network},
  url          = {https://compliancepodcastnetwork.net/stepping-up-and-stepping-forward-the-future-of-compliance-in-an-age-of-ai-and-deregulation/},
  note         = {[Accessed 22-10-2025]},
}

@misc{gu2025surveyllmasajudge,
      title={A Survey on {LLM}-as-a-Judge}, 
      author={Jiawei Gu and Xuhui Jiang and Zhichao Shi and Hexiang Tan and Xuehao Zhai and Chengjin Xu and Wei Li and Yinghan Shen and Shengjie Ma and Honghao Liu and Saizhuo Wang and Kun Zhang and Yuanzhuo Wang and Wen Gao and Lionel Ni and Jian Guo},
      year={2025},
      eprint={2411.15594},
      archivePrefix={arXiv},
      primaryClass={cs.CL},
      url={https://arxiv.org/abs/2411.15594}, 
}

@misc{villalobos2024rundatalimitsllm,
      title={Will we run out of data? Limits of {LLM} scaling based on human-generated data}, 
      author={Pablo Villalobos and Anson Ho and Jaime Sevilla and Tamay Besiroglu and Lennart Heim and Marius Hobbhahn},
      year={2024},
      eprint={2211.04325},
      archivePrefix={arXiv},
      primaryClass={cs.LG},
      url={https://arxiv.org/abs/2211.04325}, 
}

@inproceedings{schmidtova2024automatic,
  title={Automatic Metrics in Natural Language Generation: A Survey of Current Evaluation Practices},
  author={Schmidtova, Patr{\'\i}cia and Mahamood, Saad and Balloccu, Simone and Dusek, Ondrej and Gatt, Albert and Gkatzia, Dimitra and Howcroft, David M. and Platek, Ondrej and Sivaprasad, Adarsa},
  booktitle={Proceedings of the 17th International Natural Language Generation Conference},
  year={2024},
  address={Tokyo, Japan},
  pages={557--583},
  publisher={Association for Computational Linguistics},
  doi={10.18653/v1/2024.inlg-main.44},
  url={https://aclanthology.org/2024.inlg-main.44/}
}

@article{wu2023large,
  title={Large Language Models are Diverse Role-Players for Summarization Evaluation},
  author={Wu, Ning and Gong, Ming and Shou, Linjun and Liang, Shining and Jiang, Daxin},
  journal={arXiv preprint arXiv:2303.15078},
  year={2023},
  url={https://arxiv.org/abs/2303.15078}
}

@article{celikyilmaz2020evaluation,
  title={Evaluation of Text Generation: A Survey},
  author={Celikyilmaz, Asli and Clark, Elizabeth and Gao, Jianfeng},
  journal={arXiv preprint arXiv:2006.14799},
  year={2020},
  url={https://arxiv.org/abs/2006.14799}
}

@misc{scaramuzza2025showcomplyshowinganything,
      title={"{S}how Me You Comply... Without Showing Me Anything": Zero-Knowledge Software Auditing for {AI}-Enabled Systems}, 
      author={Filippo Scaramuzza and Renato Cordeiro Ferreira and Tomaz Maia Suller and Giovanni Quattrocchi and Damian Andrew Tamburri and Willem-Jan van den Heuvel},
      year={2025},
      eprint={2510.26576},
      archivePrefix={arXiv},
      primaryClass={cs.SE},
      url={https://arxiv.org/abs/2510.26576}, 
}

@article{Wagner2025,
  author       = {Wagner, Matthias and Song, Qunying and Borg, Markus and Engström, Emelie and Lysek, Michal},
  title        = {{AI Act} High-Risk {AI} Compliance Challenge and Industry Impact: A Multiple Case Study},
  journal      = {SSRN Electronic Journal},
  pages        = {51},
  year         = {2025},
  note         = {Available at SSRN: https://ssrn.com/abstract=5221279},
  doi          = {10.2139/ssrn.5221279}
}

@article{doi:10.1177/03400352251384915,
  author  = {Leo S. Lo},
  title   = {Artificial intelligence regulation matures: Landscapes of the {USA}, {European Union}, and {China}},
  journal = {IFLA Journal},
  year    = {2025},
  note    = {First published online 21 October 2025},
  doi     = {10.1177/03400352251384915},
  url     = {https://doi.org/10.1177/03400352251384915}
}

@online{codersstop2025_inconvenient_truth,
  author       = {{Coders Stop}},
  title        = {The Inconvenient Truth About {AI} Training Data That Companies Are Hiding},
  year         = {2025},
  month        = jul,
  day          = {5},
  journal      = {Medium},
  url          = {https://medium.com/@coders.stop/the-inconvenient-truth-about-ai-training-data-that-companies-are-hiding-1a3545993164}
}

@online{krasadakis2023_regulate_ai,
  author       = {George Krasadakis},
  title        = {To Regulate or Not? {H}ow should Governments React to the {AI} Revolution?},
  year         = {2023},
  month        = oct,
  day          = {20},
  journal      = {60 Leaders (Medium)},
  url          = {https://medium.com/60-leaders/to-regulate-or-not-how-should-governments-react-to-the-ai-revolution-c254d176304f},
  note         = {32 min read},
}

@misc{shen2025llmsscalinghitwall,
      title={Will {LLM}s Scaling Hit the Wall? {B}reaking Barriers via Distributed Resources on Massive Edge Devices}, 
      author={Tao Shen and Didi Zhu and Ziyu Zhao and Zexi Li and Chao Wu and Fei Wu},
      year={2025},
      eprint={2503.08223},
      archivePrefix={arXiv},
      primaryClass={cs.DC},
      url={https://arxiv.org/abs/2503.08223}, 
}

@inproceedings{nicenboim2022_explanations,
  author       = {Iohanna Nicenboim and Elisa Giaccardi and Johan Redstr{\"o}m},
  title        = {From Explanations to Shared Understandings of {AI}},
  booktitle    = {Proceedings of the DRS2022 International Conference: Bilbao},
  year         = {2022},
  month        = jun,
  publisher    = {Design Research Society},
  address      = {Bilbao, Spain},
  note         = {DRS Biennial Conference Series},
  url          = {https://dl.designresearchsociety.org/cgi/viewcontent.cgi?article=3091&context=drs-conference-papers}
}

@article{oreilly2025_eu_ai_embarrassment,
  author       = {O’Reilly, Thomas},
  title        = {The {EU}’s Approach to {AI} Is an Embarrassment},
  journal      = {The Critic},
  year         = {2025},
  month        = feb,
  day          = {14},
  note         = {Published in the "Artillery Row" section},
  url          = {https://thecritic.co.uk/the-eus-approach-to-ai-is-an-embarrassment/}
}

@article{siebel2024_ai_models_complex,
  author       = {Confino, Paolo},
  title        = {{Tom Siebel}: {AI} models are too complex for regulators—new government agencies won’t help},
  journal      = {Yahoo Finance},
  year         = {2024},
  month        = sep,
  day          = {20},
  note         = {Interview on regulatory challenges concerning AI model complexity},
  url          = {https://finance.yahoo.com/news/tom-siebel-ai-models-too-091000461.html}
}

@article{balayn_gurses2024_misguided,
  author       = {Balayn, Agathe and Gürses, Seda},
  title        = {Misguided: {AI} regulation needs a shift in focus},
  journal      = {Internet Policy Review},
  volume       = {13},
  number       = {3},
  year         = {2024},
  month        = sep,
  day          = {30},
  note         = {Open access opinion piece},
  url          = {https://policyreview.info/articles/news/misguided-ai-regulation-needs-shift/1796}
}

@article{reuters2025openai_cheapest_chatgpt,
  author       = {Reuters},
  title        = {OpenAI rolls out cheapest {ChatGPT} plan at \$4.6 in {I}ndia to chase growth},
  journal      = {Reuters},
  year         = {2025},
  month        = aug,
  day          = {19},
  note         = {Updated August 19, 2025} ,
  url          = {https://www.reuters.com/world/india/openai-rolls-out-cheapest-chatgpt-plan-46-india-chase-growth-2025-08-19/}
}

@misc{schnabl2025attestableauditsverifiableai,
      title={Attestable Audits: Verifiable {AI} Safety Benchmarks Using Trusted Execution Environments}, 
      author={Christoph Schnabl and Daniel Hugenroth and Bill Marino and Alastair R. Beresford},
      year={2025},
      eprint={2506.23706},
      archivePrefix={arXiv},
      primaryClass={cs.AI},
      url={https://arxiv.org/abs/2506.23706}, 
}

@article{addey2023charting,
  author       = {Addey, Mark},
  title        = {Charting a New Era: the {E}uropean {U}nion's {AI} Legislation and Its Transformative Influence on Technology and Society},
  year         = {2023},
  journal      = {SSRN Electronic Journal},
  doi          = {10.2139/ssrn.4560262},
  url          = {https://ssrn.com/abstract=4560262}
}

@ARTICLE{10444954,
  author={Wang, Liyuan and Zhang, Xingxing and Su, Hang and Zhu, Jun},
  journal={IEEE Transactions on Pattern Analysis and Machine Intelligence}, 
  title={A Comprehensive Survey of Continual Learning: Theory, Method and Application}, 
  year={2024},
  volume={46},
  number={8},
  pages={5362-5383},
  doi={10.1109/TPAMI.2024.3367329}}

@article{sovrano2025simplifying,
  title={Simplifying software compliance: {AI} technologies in drafting technical documentation for the {AI} Act},
  author={Sovrano, F. and Hine, E. and Anzolut, S. and others},
  journal={Empirical Software Engineering},
  volume={30},
  number={91},
  year={2025},
  doi={10.1007/s10664-025-10645-x}
}

@misc{marino2025bridgegapsmachineunlearning,
      title={Bridge the Gaps between Machine Unlearning and {AI} Regulation}, 
      author={Bill Marino and Meghdad Kurmanji and Nicholas D. Lane},
      year={2025},
      eprint={2502.12430},
      archivePrefix={arXiv},
      primaryClass={cs.LG},
      url={https://arxiv.org/abs/2502.12430}, 
}

@article{chuck_close_esq0102,
  author       = {Andy Ward},
  title        = {What {I}’ve Learned: Chuck {C}lose},
  journal      = {Esquire},
  year         = {2007},
  url          = {https://www.esquire.com/entertainment/interviews/a2048/esq0102-jan-close/},
  accessdate   = {Accessed: [Accessed 05-08-2025]}
}
}


\appendix

\end{document}